\documentclass{article}
\usepackage{nips10submit_e,times}

\usepackage{amsmath,amssymb,amsthm,graphicx}
\usepackage{subfigure}
\usepackage{wrapfig}
\usepackage{float} 
\usepackage{bm}
\usepackage{appendix}
\usepackage{url}
\usepackage{picture}

\newcommand{\zv}{\mathbf{z}}
\newcommand{\cv}{\mathbf{c}}

\newcommand{\Bv}{\mathbf{B}}
\newcommand{\Ev}{\mathbf{E}}

\usepackage{color}

\title{Infinite Hierarchical MMSB Model for Nested Communities/Groups in Social Networks}

\author{
Qirong Ho\\
School of Computer Science\\
Carnegie Mellon University\\
Pittsburgh, PA 15213\\
\texttt{qho@cs.cmu.edu}\\
\And
Ankur P. Parikh\\
School of Computer Science\\
Carnegie Mellon University\\
Pittsburgh, PA 15213\\
\texttt{apparikh@cs.cmu.edu}\\
\And
Le Song\\
School of Computer Science\\
Carnegie Mellon University\\
Pittsburgh, PA 15213\\
\texttt{lesong@cs.cmu.edu}\\
\And
Eric P. Xing\\
School of Computer Science\\
Carnegie Mellon University\\
Pittsburgh, PA 15213\\
\texttt{exping@cs.cmu.edu}\\
}

\nipsfinalcopy 

\begin{document}

\maketitle

\begin{abstract}
Actors in realistic social networks play not one but a number of diverse roles depending on whom they interact with, and a large number of such role-specific interactions collectively determine social communities and their organizations. 
Methods for analyzing social networks should capture these multi-faceted role-specific interactions, and, more interestingly, discover the latent organization or hierarchy of social communities. We propose a hierarchical Mixed Membership Stochastic Blockmodel to model the generation of hierarchies in social communities, selective membership of actors to subsets of these communities, and the resultant networks due to within- and cross-community interactions. Furthermore, to automatically discover these latent structures from social networks, we develop a Gibbs sampling algorithm for our model. We conduct extensive validation of our model using synthetic networks, and demonstrate the utility of our model in real-world datasets such as predator-prey networks and citation networks.

\end{abstract}

\section{Introduction}

How do the social communities and their self-organization arise from coordinated interactions and information sharing among the actors? One way to tap into this question is to understand the latent roles and minds of actors which lead to the formation and organization of these communities. We are particularly interested in uncovering the functional/sociological underpinning of network actors, and their influence on the network modularity and hierarchy. 

Existing methods for inferring actor roles~\cite{Hoff:Raft:Hand:2002,McCallum_ICML06,handcockRSS06} are limited in that they assume each actor undertakes a single, invariant role when interacting with other actors. This is in stark constrast with real social networks where actors can play multiple roles (or be under multiple cultural influences); the specific role being played depends on whom the actor is interacting with. The {\it Mixed membership stochastic blockmodel} (MMSB)~\cite{airoldi2008mixed} proposed recently captures the multi-faceted, context-specific nature of an actor's role.

\begin{wrapfigure}{r}{0.53\textwidth}

    \begin{center}
        \includegraphics[width=0.53\textwidth]{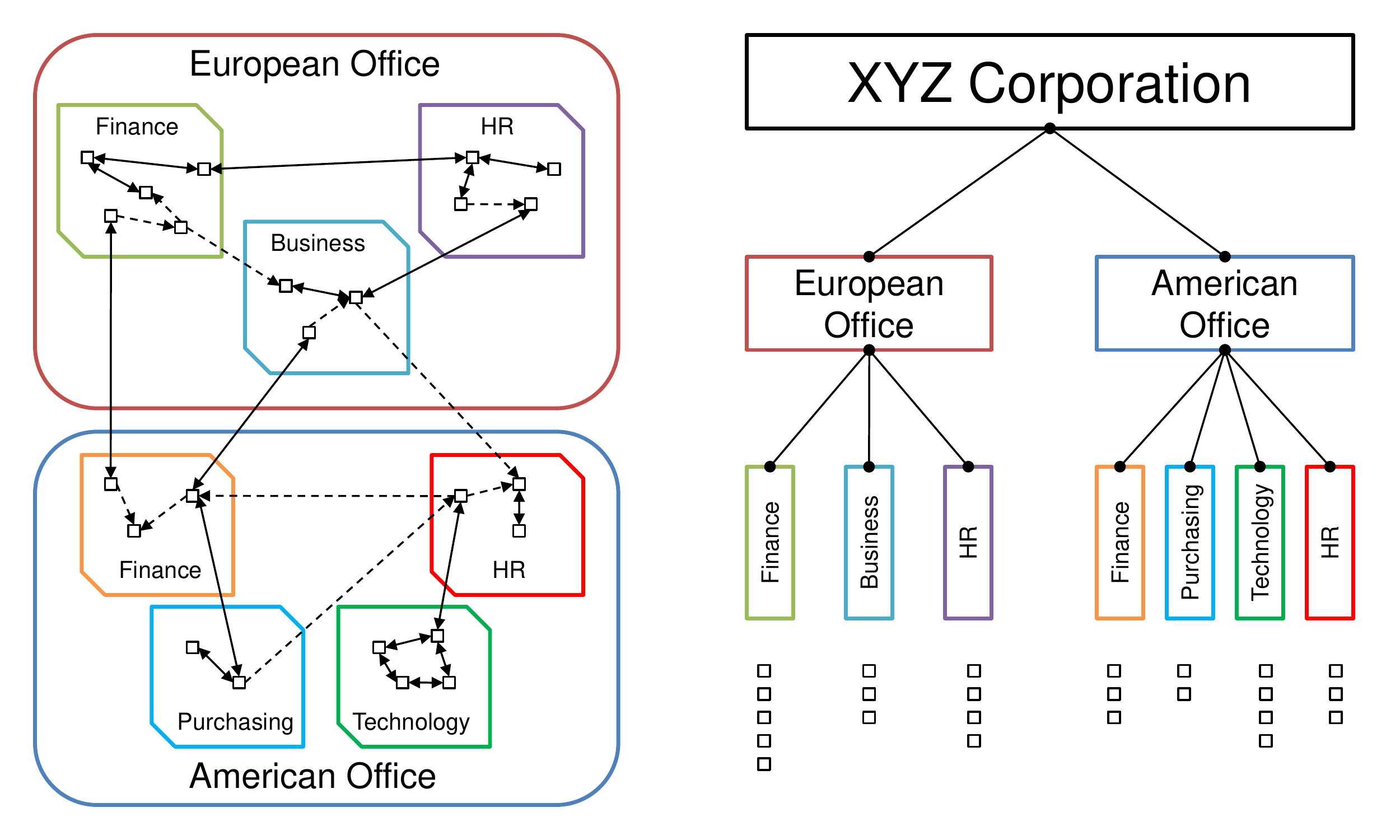}
    \end{center}

    \caption{\footnotesize Illustration of a corporate email network/graph as a set of nested communities {\bf (Left)},
        and the corresponding hierarchy of communities {\bf (Right)}.
        Vertices in the network represent employees, and directed edges represent emails sent on a particular day.}
    \label{fig:community_illustration}

\end{wrapfigure}

However, the presence of hierarchical structures in real-world social networks raises a significant challenge to current network models. The MMSB formalism described above is intrinsically flat in its structure --- all roles of an actor are equal citizens in terms of their relationships to each other --- so it does not induce hierarchical structures among actors. A hierarchical structure goes beyond simple clustering by explicitly including organization at all scales in a network simultaneously. Taking a corporate email network as example (Figure \ref{fig:community_illustration}), suppose a finance department employee in the European branch is communicating with an American branch executive. The European employee might not communicate with a (possibly business-oriented) role related to her membership in the European finance department (which normally has no business with the American finance department), but simply with a more generic social role as a member of the European branch of the same company (which has occasional acquaintance with the American colleagues). However, when this same employee interacts with a European human resource employee, she would likely do so with her role as a European finance department member. This notion of {\it interaction-specific, multi-community membership} accounts for otherwise unusual edges --- like the interaction between the European finance employee and the American executive --- and motivates an intuitive yet powerful approach that allows one to infer and visualize the semantic underpinnings of every actor and every link between actors.

We propose a tree-structured hierarchical MMSB (hMMSB) to model community-hierarchies and the resultant social network as a function of the community memberships undertaken by actors during different interactions. Our method treats the community structure as a hierarchy of super- and sub-communities, while accounting for the interaction-specific nature of community membership. Moreover, it automatically determines the appropriate number of sub-communities in each community. This sets our method apart from spectral clustering methods, which typically do not produce hierarchies, or agglomerative clustering methods, which are inherently restricted to binary hierarchies. We will use our model to analyze a biological food web and a physics paper citation network, and examine whether the recovered hierarchies agree with the known organizational structure of both networks. Extensive simulation-based validation of both our model's consistency and the Gibbs-sampling inference algorithm will be conducted to ensure our model's feasibility for hierarchical structure identification.

\subsection{Related Work}

While there has been a great deal of work on graph clustering and community structure inference~\cite{girvan:newman:2002,krause:frank:mason:ulanowicz:taylor:2003,radicchi:castellano:cecconi:loreto:parisi:2004,clauset:newman:moore:2004,guimera:amaral:2005}, probabilistic generative models for hierarchical community formation and latent multi-role inference of every actor and link have just started to draw attention.

As mentioned earlier, the MMSB model~\cite{airoldi2008mixed} enables inference of the latent roles of every actor and link in a network, but it cannot capture hierarchical structures of possible communities in the network. 
The link prediction model in \cite{miller2009linkprediction} employs an Indian Buffet Process prior over actor positions in an infinite-dimensional latent feature space. In that respect, it may be thought of as a nonparametric extension of the MMSB. However, the goal of their model is missing link prediction rather than inference of latent organizational structure. 

Recently, Roy {\it et. al.} have developed an ``annotated hierarchies" model~\cite{roy2007learning} that generalizes an infinite relational model~\cite{tenenbaum:irm:2006} for hierarchical group discovery. Their subsequent work on Mondrian Processes~\cite{roy:mondrian:2009} provides an alternative nonparametric take on the same issues. Both the annotated hierarchies model and the Mondrian Process relational model discover {\it binary} hierarchies. 
The infinite stochastic blockmodel~\cite{kemp:pnas:2008} recovers organizational hierarchies of arbitrary branching factor, among other types of structure like grids or partitions, but it does not offer detailed semantic underpinnings, such as context-dependent role or community membership instantiations, of every network link.




\section{Probabilistic Hierarchical Community Model}


Under the MMSB~\cite{airoldi2008mixed}, network links can be instantiated by a role-specific {\em local} interaction mechanism: the link between each pair of actors, say $(i,j)$, is instantiated according to the latent role specifically undertaken by actor $i$ when it is to interact with $j$, and that of $j$ when it is to interact with $i$. Crucial to the goal of role-prediction for network data, is the so-called ``role vector'' $\theta$ of {\it mixed membership coefficients}, which succinctly captures the probabilities of an actor being involved in different roles when he/she interacts with another actor. In this paper, we leverage and generalize the ``role vector'' to model context-dependent mixed memberships in communities of different scales when links between actors are formed. We propose a {\it hierarchical MMSB} (hMMSB) model for nested community structures underlying social networks, thereupon such structures as well as the community implications of every link can be inferred, given the network.  
In the sequel, we begin with a baseline hMMSB that requires a user to pre-specify the granularity of a tree hierarchy of communities, that is, the branching degree from each group to its subgroup at the next level. We will then relax this constraint by employing a nested Chinese Restaurant Process, described in ~\cite{blei2010nested}, to allow hierarchies of arbitrary granularities with a fixed depth.

Let $G=\{V,E\}$ denote a directed graph, where $V=\{1,\dots,N\}$ is the set of vertices/actors, and $E$ is the set of edges/interactions. We adopt the convention $E_{ij}=1$ if $(i,j)\in E$ else $E_{ij}=0$, and we ignore self-edges $E_{ii}$. Because the graph is directed, $E$ is not necessarily a symmetric matrix. Given an edge $E_{ij}$, we refer to actor $i$ as the {\it donor} and actor $j$ as the {\it receiver}.

Under an hMMSB, we assume that a link $E_{ij}$ follows a generative process based on two latent variables unique to each actor. The first one is a {\it community membership-path vector} (or in short, {\it path}) $c_i$, which records the community memberships of this actor at different levels of the community hierarchy in the network. 
Our model requires an integer parameter $K$, denoting the maximum depth to which the hierarchy should be learnt. Given $K$, we can represent an actor path $c_i$ as a vector of {\it positive integers} $(c_{i1},c_{i2},\dots,c_{iK})$, where $c_{ik}$ denotes the branch taken by the path at level $k\ge 1$. We adopt the convention that the hierarchy root sits at level 0. For clarity of explanation, we will assume for now that the actor paths $c_i$ are known. Later, we shall relax this assumption by placing a suitable nonparametric prior over $c_i$, allowing them to be posteriorly inferred. 

The second variable is a {\it mixed membership vector} (in short, {\it MM}) $\theta_i$, defining the probabilities of an actor identifying himself at different community levels when interacting with other actors. We place a two-parameter stick-breaking prior \cite{blei2010nested} over $\theta_i$, so that actors are not required to participate in the same number of community levels. Stick breaking constructions work as follows: Consider a stick of length 1. Draw  $V_{i1} \sim Beta(m\pi, (1-m)\pi)$. Let $\theta_{i1} = V_{i1}$ and let $1-\theta_{i1}$ be the remainder of the stick after chopping off this length $V_{i1}$. To calculate the length $\theta_{i2}$, draw $V_{i2} \sim Beta(m\pi, (1-m)\pi)$ and chop off this fraction of the remainder of the stick, giving $\theta_{i2} = V_{i2} (1-V_{i1})$. Thus $V_{ik}$ is the fraction to chop off from the stick's remainder, and $\theta_{ik}$ is the length of the $kth$ stick that was chopped off.
In general, we draw $V_{ik} \sim Beta(m\pi, (1-m)\pi)$ from $k=1$ to $k=\infty$  and the corresponding $\{\theta_{ik}\}_{k=1}^{\infty}$ is defined below:
\begin{equation}
\theta_{ik} = V_{ik} \prod_{u=1}^{k-1} (1 - V_{iu})
\end{equation}
 This process is known as the two-parameter GEM distribution \cite{blei2010nested} and draws from $GEM(m, \pi)$ are denoted as $\theta_i \sim\mathrm{GEM}(m,\pi)$. $m>0$ influences the mean of $\theta_i$, and $\pi>0$ influences its variance. Because the hierarchy is only learnt up to depth $K$, we truncate the $\mathrm{GEM}(m,\pi)$ distribution at level $K$. The stick breaking prior makes it more intuitive to bias interactions toward coarser or finer levels compared to a Dirichlet prior with either a single parameter (which is not expressive enough), or $K-1$ parameters (which may be too expressive).
 
To instantiate the edge from actor $i$ to $j$, we introduce {\it interaction level indicators} $z_{i\rightarrow j}$ ({\it donor level}) and $z_{i\leftarrow j}$ ({\it receiver level}), which follow multinomial distributions defined by the MM vectors $\theta_i,\theta_j$ respectively. The pair $(c_i, z_{i\rightarrow j})$ specifies actor $i$'s interaction-specific position in the community hierarchy, denoted by $c_i[z_{i\rightarrow j}]$, which contributes to determining his interaction probability with actor $j$. Note that we are {\it not} proposing a full hierarchical model for a tree over conditionally $iid$ entities, as in a coalescent process~\cite{hudson1990coalescent}. Our goal is to model finite-depth nested communities of entities connected by a network; therefore our latent hierarchy is not a full binary tree over all nodes, and our generative process does not output $iid$ nodal attributes, but links between nodes. 

Now, given interaction level indicators $z_{i\rightarrow j}$ and $z_{i\leftarrow j}$, and paths $c_i$ and $c_j$ of the two actors involved, we can determine the community identities (not necessarily at the same level) underlying an interaction. These community identities specify a community-pair-specific distribution for the link $E_{ij}\sim\mathrm{Bernoulli}(\mathrm{S}_B(c_i,c_j,z_{i\rightarrow j},z_{i\leftarrow j}))$, where
\begin{eqnarray}
	\mathrm{S}_B(c_i,c_j,z_{i\rightarrow j},z_{i\leftarrow j}) &=&
		\begin{cases}
		B_{c_i,c_j,z_{\mathrm{coarse}}} & (c_{i,1},\dots,c_{i,z_{\mathrm{coarse}}-1}) = (c_{j,1},\dots,c_{j,z_{\mathrm{coarse}}-1}) \\
		0 & \mathrm{otherwise}
		\end{cases}
		\label {eq:f_def} \\
	z_{\mathrm{coarse}} &=& \min(z_{i\rightarrow j},z_{i\leftarrow j}). \notag
\end{eqnarray}
The notation $B_{c_i,c_j,z_{\mathrm{coarse}}}\in \Bv$ denotes a parameter specifying the interaction probability from community $c_{i}[z_{\mathrm{coarse}}]$ to $c_{j}[z_{\mathrm{coarse}}]$. We shall explain the intuition behind this in more detail below.

With all these details, we arrive at the following generative process of a network $G=\{V,E\}$:
{\small
\begin{itemize}
\item
For each actor $i\in V$:
	\begin{itemize}
	\item
	Sample actor $i$'s path $c_i$ from a distribution over hierarchies (detailed in the next section). 
	\item
	Sample actor $i$'s MM $\theta_i$ (distribution over community levels): $\theta_i \sim \mathrm{GEM}(m,\pi)$.
	\end{itemize}
\item
For each element $B_{\cdot,\cdot,\cdot}$ of $\Bv$ in the community-compatability matrix:
	\begin{itemize}
	\item
	Sample a value for the element $B_{\cdot,\cdot,\cdot} \sim \mathrm{Beta}(\lambda_1,\lambda_2)$.
	\end{itemize}
\item
To generate the network, for each directed edge $E_{ij}$ where $i,j\in V$, $i\ne j$:
	\begin{itemize}
	\item
	Sample the donor level $z_{i\rightarrow j} \sim \mathrm{Multinomial}(\theta_i)$.
	\item
	Sample the receiver level $z_{i\leftarrow j} \sim \mathrm{Multinomial}(\theta_j)$.
	\item
	Sample the interaction $E_{ij}\sim\mathrm{Bernoulli}(\mathrm{S}_B(c_i,c_j,z_{i\rightarrow j},z_{i\leftarrow j}))$.
	\end{itemize}
\end{itemize}}

The overall intuition behind hMMSB is that $c_i$ and $z_{i\rightarrow j}$ specify a hierarchy position for the donor, and likewise for the receiver; and we use these positions to determine the interaction probability. There are two different scenarios worth mentioning. First, suppose that the donor and receiver positions share the same immediate parent in the hierarchy. Our model associates a ``compatibility matrix" with every parent position in the hierarchy, which gives the probability of interaction between any two of its child positions. The set of all compatibility matrices is denoted by $\Bv$. Since the donor and receiver positions share the same parent, we simply look up the appropriate entry in $\Bv$ to get the interaction probability.

\begin{wrapfigure}{r}{0.5\textwidth}
	\begin{center}
		\includegraphics[width=0.5\textwidth]{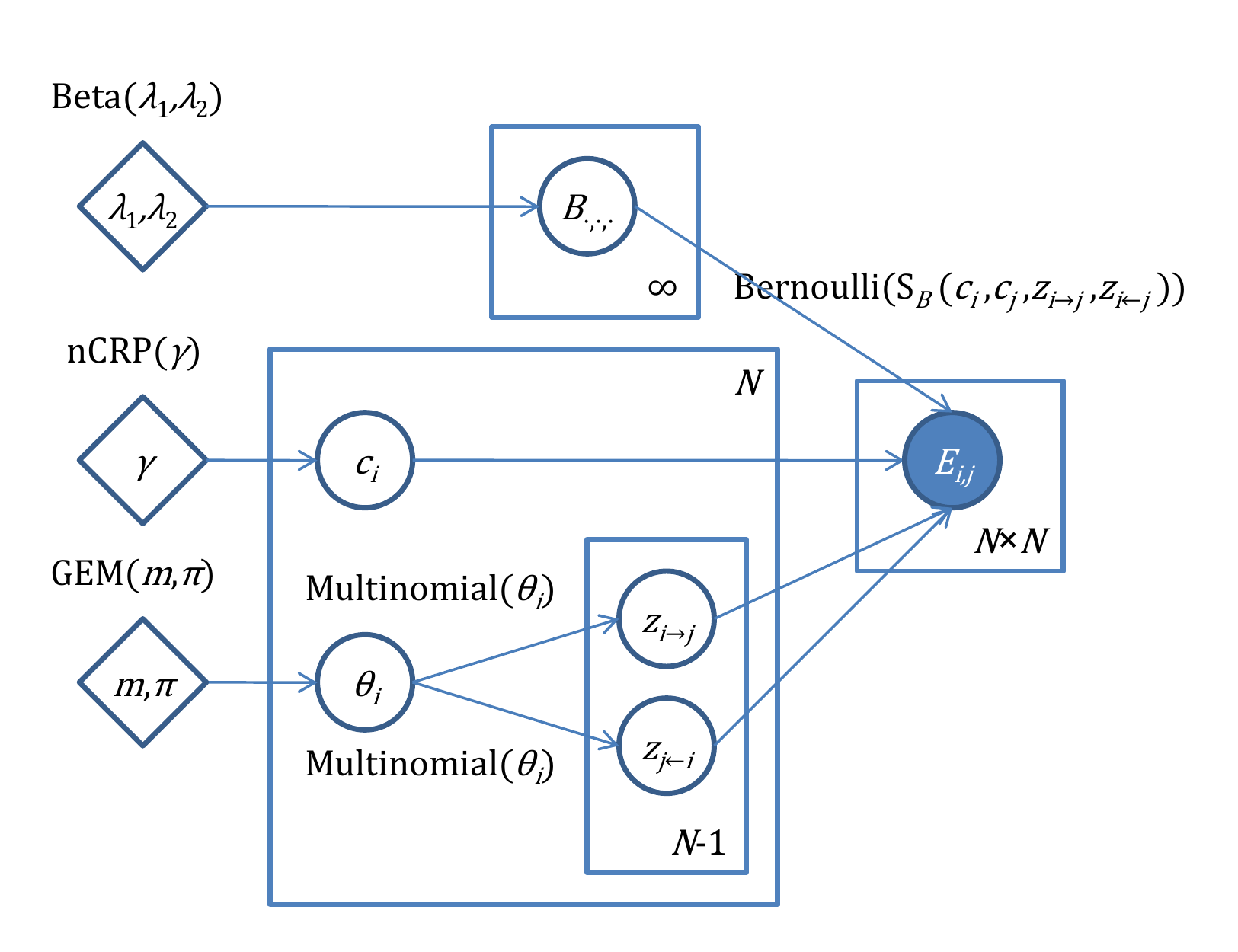}
	\end{center}
	\caption{\footnotesize Graphical model for hMMSB.}
	\label{fig:graphical_model}
\end{wrapfigure}

In the second case, the donor and receiver positions do not share the same parent; this always happens when $z_{i\rightarrow j}\ne z_{i\leftarrow j}$ (i.e. the positions are at different depths). We then coarsen both levels to their minimum, $z_{\mathrm{coarse}}$. The idea is that interactions between two hierarchy positions should take place at the level of the higher (coarser) position. For example, when an executive interacts with a sales representative, we reduce the interaction to one between an executive and a generic group of ``other employees'' below executive rank. (It is in theory possible to explicitly model all possible interactions between communities of arbitrary levels, but this will lead to a severe over-parameterization of our model. We have decided to postpone a more thorough study of this issue to later work.) If the new positions now share the same parent, then we look up the appropriate element of $\Bv$ as before. Otherwise, we define the interaction probability to be zero. 





\subsection{Infinite hMMSB}

Thus far, we have assumed knowledge of the paths $c_i$. From a combinatorial standpoint, inferring these paths is difficult. Because the number of children at each hierarchy position is unlimited, there are infinitely many possible hierarchies or sets of paths $\{c_i\}$, even with the fixed depth $K$. One might consider using heuristic methods that guess at the number of such children, but doing so would defeat the purpose of employing a probabilistic model.

Our solution is to place an appropriate nonparametric Bayesian prior over $c_i$, the Nested Chinese Restaurant Process \cite{blei2010nested}, or nCRP for short. The nCRP is an extension of the regular Chinese Restaurant Process (CRP), a recursively-defined prior over positive integers. For concreteness, we shall use the first level of each actor path, $c_{i1}$, to define the CRP:
\begin{eqnarray}
	\mathrm{P}(c_{i1}=x \mid c_{1:(i-1),1}) =
	\begin{cases}
		\frac{\left|\{j<i \,|\, c_{j1}=x\}\right|}{i-1+\gamma} & x\in\{c_{1:(i-1),1} \} \\
		\frac{\gamma}{i-1+\gamma} & \text{$x$ is the smallest positive integer not in $\{c_{1:(i-1),1}\}$}
	\end{cases}
\end{eqnarray}
where $\gamma>0$ is a ``concentration'' parameter that controls the probability of drawing new integers, and for conciseness we define $c_{1:(i-1),1} \equiv (c_{11},\dots,c_{(i-1)1})$. The nCRP is essentially a hierarchy of CRP priors, beginning with a single CRP prior at the top level. With each unique integer $x$ seen at the top-level prior, we associate a child CRP prior with $|\{i\,|\,c_{i1}=x\}|$ observations, resulting in a two-level tree of CRP priors. We can repeat this process {\it ad infinitum} on the newly-created child priors, resulting in an infinte-level tree of CRP priors, though we only use a $K$-level nCRP. All CRP priors in the nCRP share the same concentration parameter $\gamma$.

Now we can finish describing our generative process: for each actor $i\in V$, we can sample $c_i\sim\mathrm{nCRP}(\gamma)$ using the recursive nCRP definition:
\begin{eqnarray}
	&\hspace{-0.8cm}&\mathrm{P}(c_{ik}=x \mid c_{1:(i-1)}, c_{i,1:(k-1)}) = \notag \\
	&\hspace{-0.8cm}&\begin{cases}
		\frac{\left|\{ j<i \,|\, c_{j,1:(k-1)}=c_{i,1:(k-1)} \wedge c_{jk}=x \}\right|}
			{\left|\{ j<i \,|\, c_{j,1:(k-1)}=c_{i,1:(k-1)} \}\right| + \gamma}
		& x \in \{c_{jk} \,|\, (j<i) \wedge c_{j,1:(k-1)}=c_{i,1:(k-1)} \} \\
		\frac{\gamma}
			{\left|\{ j<i \,|\, c_{j,1:(k-1)}=c_{i,1:(k-1)} \}\right| + \gamma}
		& \text{$x$ is the smallest positive integer not in the above set.}
	\end{cases}
	\label{eq:nCRP_defn}
\end{eqnarray}
Figure \ref{fig:graphical_model} displays our complete generative process as a graphical model. As a side note, the infinite nCRP prior on paths implies that $\Bv$ contains an infinite number of elements (there could be infinitely many children at each parent, so the compatibility matrices must be infinite-dimensional). Our Gibbs sampler finesses this issue by integrating out $\Bv$, while the posterior distribution of each $B_{\cdot,\cdot,\cdot}$ can be recovered from the path and level samples.

\section{Collapsed Gibbs Sampler}

Exact inference on our model is intractable, so we derive a collapsed Gibbs sampling scheme for posterior inference. The $\theta$'s and $\Bv$'s are integrated out for faster mixing, so we only have to sample $\zv$ and $\cv$. The sampling equations are provided below.


\paragraph{Sampling levels}

The distribution of $z_{i\rightarrow j}$ conditioned on all other variables is
\begin{eqnarray}
&& \mathbb{P} (z_{i\rightarrow j}  \mid \mathbf{c}, \mathbf{z}_{-(i \rightarrow j)}, \mathbf{E}, \gamma, m, \pi, \lambda_1, \lambda_2) \notag \\
	&\propto& \mathbb{P} (E_{i,j}, z_{i\rightarrow j}  \mid \mathbf{c}, \mathbf{z}_{-(i \rightarrow j)},
		\mathbf{E}_{-(i,j)}, \gamma, m, \pi, \lambda_1, \lambda_2) \notag \\
	&=& \mathbb{P} (E_{i,j} \mid \mathbf{c},  \mathbf{z}, \mathbf{E}_{-(i,j)}, \gamma, m, \pi, \lambda_1, \lambda_2)
		\mathbb{P} (z_{i\rightarrow j}  \mid \mathbf{c},  \mathbf{z}_{-(i \leftarrow j)},
		\mathbf{E}_{-(i,j)}, \gamma, m, \pi, \lambda_1, \lambda_2) \notag \\
	&=& \mathbb{P} (E_{i,j} \mid \cv, \zv, \Ev_{-(i,j)}, \lambda_1, \lambda_2)
		\mathbb{P} (z_{i\rightarrow j} \mid \mathbf{z}_{i,(-j)}, m, \pi) \label{eq:sample_z} 
\end{eqnarray}
where $\mathbf{E}_{-(i,j)}$ is the set of all edges except $E_{ij}$, and $\mathbf{z}_{i,(-j)} = \{z_{i\rightarrow\cdot},z_{\cdot\leftarrow i}\} \setminus z_{i\rightarrow j}$. By Beta-Bernoulli conjugacy, the first term, for a particular value of $z_{i\rightarrow j}$, is
\begin{eqnarray}
\text{First term} &=&
	\begin{cases}
		\frac{E_{ij}(a+\lambda_1)+(1-E_{ij})(b+\lambda_2)}{a+b+\lambda_1+\lambda_2}
			& \mathrm{S}_B(E_{ij}) \ne 0 \\
		0 & \text{otherwise}
	\end{cases} \notag \\
a &=& \left|\left\{
	(x,y) \mid (x,y)\ne(i,j) \,\wedge\,
	\mathrm{S}_B(E_{xy}) = \mathrm{S}_B(E_{ij}) \,\wedge\,
	E_{xy} = 1
	\right\}\right| \notag \\
b &=& \left|\left\{
	(x,y) \mid (x,y)\ne(i,j) \,\wedge\,
	\mathrm{S}_B(E_{xy}) = \mathrm{S}_B(E_{ij}) \,\wedge\,
	E_{xy} = 0
	\right\}\right|
\label{eq:likelihood}
\end{eqnarray}
where we have defined the shorthand $\mathrm{S}_B(E_{ij}) \equiv \mathrm{S}_B(c_i,c_j,z_{i\rightarrow j},z_{i\leftarrow j})$. 

The second term can be computed by iterated expectation, conditioning on actor $i$'s
stick breaking lengths $V_{i1}$,...,$V_{ik}$:
\begin{eqnarray}
	&& \mathbb{P} (z_{i\rightarrow j}=k \mid \mathbf{z}_{i,(-j)}, m, \pi) \notag \\
	&=& \mathbb{E} \left[ \mathbb{I}(z_{i\rightarrow j}=k) \mid \mathbf{z}_{i,(-j)}, m, \pi \right] \notag \\
	&=& \mathbb{E} \left [ \mathbb{E} \left[ \mathbb{I}(z_{i\rightarrow j}=k) \mid
		V_{i1},...,V_{ik}, \mathbf{z}_{i,(-j)}, m, \pi \right ] \right] \notag \\
	&=& \mathbb{E} \left [ V_{ik} \prod_{u=1}^{k-1} (1 - V_{iu}) \mid \mathbf{z}_{i,(-j)}, m, \pi \right] \notag \\
	&=& \mathbb{E}[V_{ik} \mid \mathbf{z}_{i,(-j)}, m, \pi]
		\prod_{u=1}^{k-1} \mathbb{E}[(1 - V_{iu}) \mid \mathbf{z}_{i,(-j)}, m, \pi] \notag \\
	&=& \frac{m\pi + \#[\mathbf{z}_{i,(-j)} = k]}{\pi + \#[\mathbf{z}_{i,(-j)} \ge k]}
		\prod_{u=1}^{k-1} \frac{(1-m)\pi +\#[\mathbf{z}_{i,(-j)} > u]}{\pi + \#[\mathbf{z}_{i,(-j)} \ge u]}
\end{eqnarray}
Since we have limited the maximum depth to $K$, we simply ignore the event $z_{i\rightarrow j} > K$, and renormalize the distribution of $z_{i\rightarrow j}$ over the domain $\{1,\dots,K\}$. The conditional distribution of $z_{i\leftarrow j}$ is derived in similar fashion.

\paragraph{Sampling paths}

The distribution of $c_i$ conditioned on all other variables is
\begin{eqnarray}
&& \mathbb{P}(c_i \mid \cv_{-i}, \zv, \Ev, \gamma, m, \pi, \lambda_1, \lambda_2) \notag \\
&\propto& \mathbb{P}(c_i, \Ev_{(i, \cdot), (\cdot, i)} \mid
	\mathbf{c}_{-i}, \mathbf{z}, \Ev_{-(i, \cdot), -(\cdot, i)}, \gamma, m, \pi, \lambda_1, \lambda_2) \notag \\
&=& \mathbb{P}(\Ev_{(i, \cdot), (\cdot, i)} \mid \cv, \zv, \Ev_{-(i, \cdot), -(\cdot, i)}, \gamma, m, \pi, \lambda_1, \lambda_2)
	\mathbb{P}(c_i \mid \cv_{-i}, \zv, \Ev_{-(i,\cdot), -(\cdot, i)}, \gamma, m, \pi, \lambda_1, \lambda_2) \notag \\
&=& \mathbb{P}(\Ev_{(i, \cdot), (\cdot, i)} \mid \cv, \zv, \Ev_{-(i, \cdot), -(\cdot, i)}, \lambda_1, \lambda_2)
	\mathbb{P}(c_i \mid \cv_{-i}, \gamma)
\label{eq:path}
\end{eqnarray}
where $\Ev_{(i, \cdot), (\cdot, i)} = \{ E_{xy} \mid x=i \,\vee\, y=i\}$ is the set of all edges $E_{ij}$ whose distributions depend on $c_i$, and $\Ev_{-(i, \cdot), -(\cdot, i)}$ is its complement. The second term can be computed using the recursive nCRP definition:
\begin{eqnarray}
	&\hspace{-0.8cm}&\mathrm{P}(c_{ik}=x \mid c_{1:(i-1)}, c_{i,1:(k-1)},\gamma) = \notag \\
	&\hspace{-0.8cm}&\begin{cases}
		\frac{\left|\{ j<i \,|\, c_{j,1:(k-1)}=c_{i,1:(k-1)} \wedge c_{jk}=x \}\right|}
			{\left|\{ j<i \,|\, c_{j,1:(k-1)}=c_{i,1:(k-1)} \}\right| + \gamma}
		& x \in \{c_{jk} \,|\, (j<i) \wedge c_{j,1:(k-1)}=c_{i,1:(k-1)} \} \\
		\frac{\gamma}
			{\left|\{ j<i \,|\, c_{j,1:(k-1)}=c_{i,1:(k-1)} \}\right| + \gamma}
		& \text{$x$ is the smallest positive integer not in the above set.}
	\end{cases}
	\label{eq:nCRP_defn}
\end{eqnarray}
while the first term, for a particular value of $c_i$, is
\begin{eqnarray}
\text{First term} \!\!\!\!\!&=&\!\!\!\!\!
	\begin{cases}
		{\displaystyle \prod_{B \in \Bv_{(i, \cdot), (\cdot, i)}} } \!\!\!\!\!\!
		\frac{\Gamma(g_B+h_B+\lambda_1+\lambda_2)}{\Gamma(g_B+\lambda_1)\Gamma(h_B+\lambda_2)}
		\frac{\Gamma(g_B+r_B+\lambda_1)\Gamma(h_B+s_B+\lambda_2)}{\Gamma(g_B+h_B+r_B+s_B+\lambda_1+\lambda_2)}
			\!\!\!\!& \forall E_{xy} \!\in\! \Ev_{(i, \cdot), (\cdot, i)}, \,\mathrm{S}_B(E_{xy}) \!\ne\! 0 \\
		0 & \text{otherwise}
	\end{cases} \notag \\
g_B &=& \left|\left\{
	(x,y) \mid E_{xy}\in\Ev_{-(i, \cdot),-(\cdot, i)} \,\wedge\,
	\,\mathrm{S}_B(E_{xy}) = B \,\wedge\,
	E_{xy} = 1
	\right\}\right| \notag \\
h_B &=& \left|\left\{
	(x,y) \mid E_{xy}\in\Ev_{-(i, \cdot),-(\cdot, i)} \,\wedge\,
	\,\mathrm{S}_B(E_{xy}) = B \,\wedge\,
	E_{xy} = 0
	\right\}\right| \notag \\
r_B &=& \left|\left\{
	(x,y) \mid E_{xy}\in\Ev_{(i, \cdot),(\cdot, i)} \,\wedge\,
	\,\mathrm{S}_B(E_{xy}) = B \,\wedge\,
	E_{xy} = 1
	\right\}\right| \notag \\
s_B &=& \left|\left\{
	(x,y) \mid E_{xy}\in\Ev_{(i, \cdot),(\cdot, i)} \,\wedge\,
	\,\mathrm{S}_B(E_{xy}) = B \,\wedge\,
	E_{xy} = 0
	\right\}\right|
\end{eqnarray}
where $\Bv_{(i, \cdot), (\cdot, i)} = \{ B\in\Bv \mid \exists (i,j), (E_{ij}\in\Ev_{(i, \cdot), (\cdot, i)} \,\wedge\, \mathrm{S}_B(E_{ij}) = B) \}$ is the set of all $B\in\Bv$ associated with some edge in $\Ev_{(i, \cdot), (\cdot, i)}$ through $\mathrm{S}_B()$.

\section{Simulation}

We evaluate our model's ability to recover hierarchies on simulated data. For all simulations, the number of actors $N$ was 150, the max depth was $K=2$ (2 levels plus root) and $\theta = (.25, .75)$ for all actors, meaning that actors interact at level 1 25\% of the time and level 2 75\% of the time. Our experiments explore the effect of different compatibility matrices $\Bv$. We first explore an on-diagonal $\Bv$, where the diagonal elements are much larger than the off-diagonal elements (actors tend to interact within their own communities). We also investigate an off-diagonal $\Bv$, where the off-diagonal elements are larger (actors tend to interact outside their own communities). In the ``low noise" setting the off-diagonal and on-diagonal elements are far apart (to clearly distinguish them), while in the ``high noise" setting they are closer together. 
The exact parameters for the 4 types of $\Bv$'s are shown below:
\begin{enumerate}
\item
\textbf{on-diagonal, low noise} - $B_{on-diagonal} = (.4, .8)$, $B_{off-diagonal} = (.02, .02)$;
\item
\textbf{on-diagonal, high noise} - $B_{on-diagonal} = (.3, .6)$, $B_{off-diagonal} = (.1, .1)$;
\item
\textbf{off-diagonal, low noise}  - $B_{on-diagonal} = (.02, .02)$, $B_{off-diagonal} = (.4, .8)$;
\item
\textbf{off-diagonal, high noise} - $B_{on-diagonal} = (.1, .1)$, $B_{off-diagonal} = (.3, .6)$.
\end{enumerate}
$B_{on-diagonal} = (a,b)$ means that actors interacting in the same level-1 community do so with probability $a$, while actors interacting in the same level-2 community do so with probability $b$ (and analogously for $B_{off-diagonal}$).

We compare our approach to hierarchical spectral clustering (denoted HSpectral). For spectral clustering, it is unclear how the number of clusters at each node would be selected, so \textit{we give it the number of 1st-level branches as an advantage} (and then let it do a binary split at each level-2 node). For hMMSB, we fix $m=\pi=\lambda_1=\lambda_2 = .5$ and search over $\gamma = \{.01, .1, .5, 1, 1.5, 2\}$, picking the value that maximizes the marginal likelihood. 

The Gibbs sampler was run for 1,500 iterations on each experiment. We calculate the F1 score at each level $k$, $\mathrm{F}1_k = \frac{2*Precision*Recall}{Recall + Precision}$ where $Recall = \frac{TP}{TP + FN}$, and $Precision =  \frac{TP}{TP + FP}$. $TP$ is true positive count (actors that are in the same cluster and should be up to depth k), $FP$ is false positive count, $TN$ is true negative count, and $FN$ is false negative count. The total F1 score is computed by averaging the $\mathrm{F}1_k$ scores for each level.

\begin{figure}

\begin{center}
 \subfigure[]{\label{fig:150ondiagonallownoise}\includegraphics[width=3.5cm]{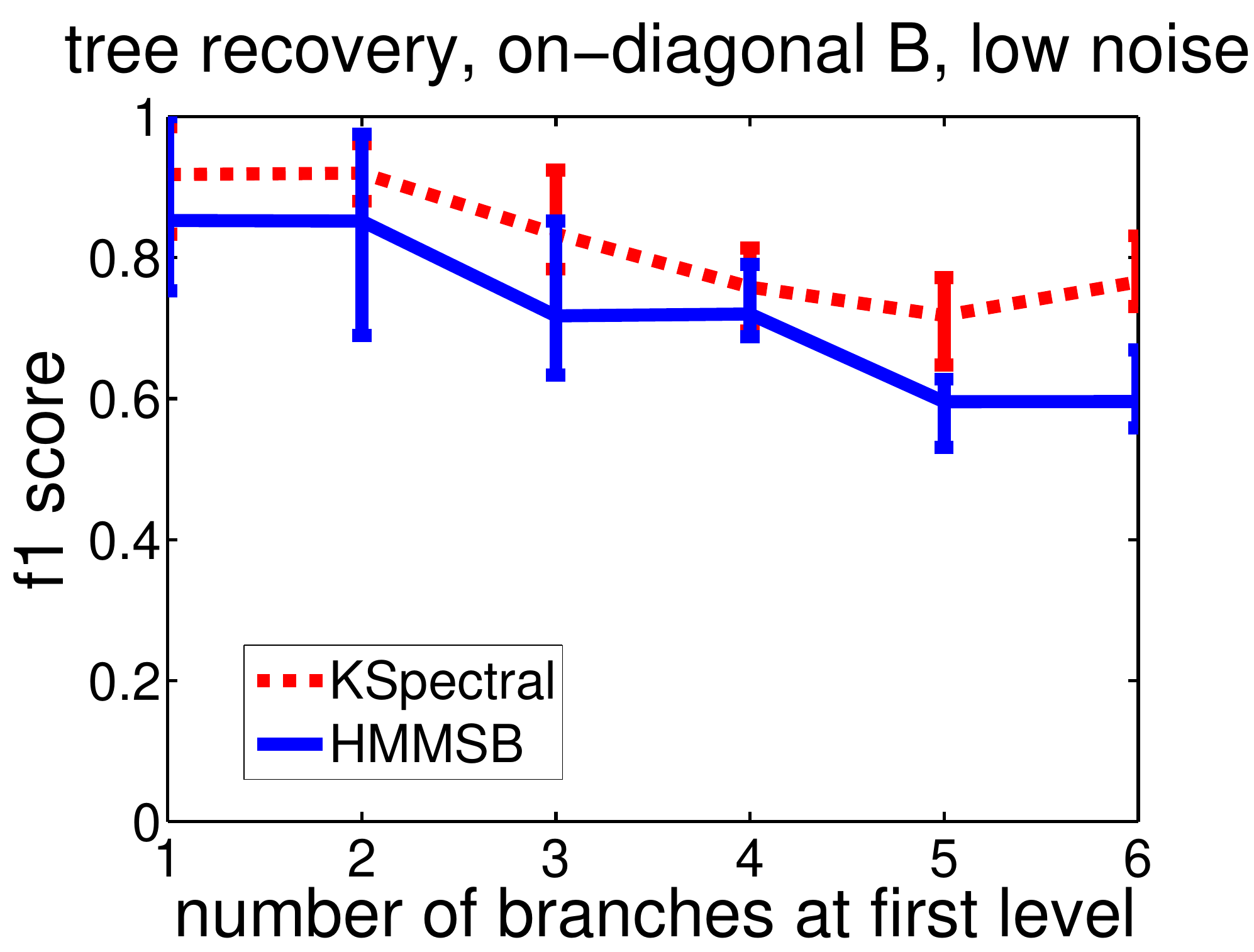}}
 \nolinebreak
 \subfigure[]{\label{fig:150ondiagonalhighnoise}\includegraphics[width=3.5cm]{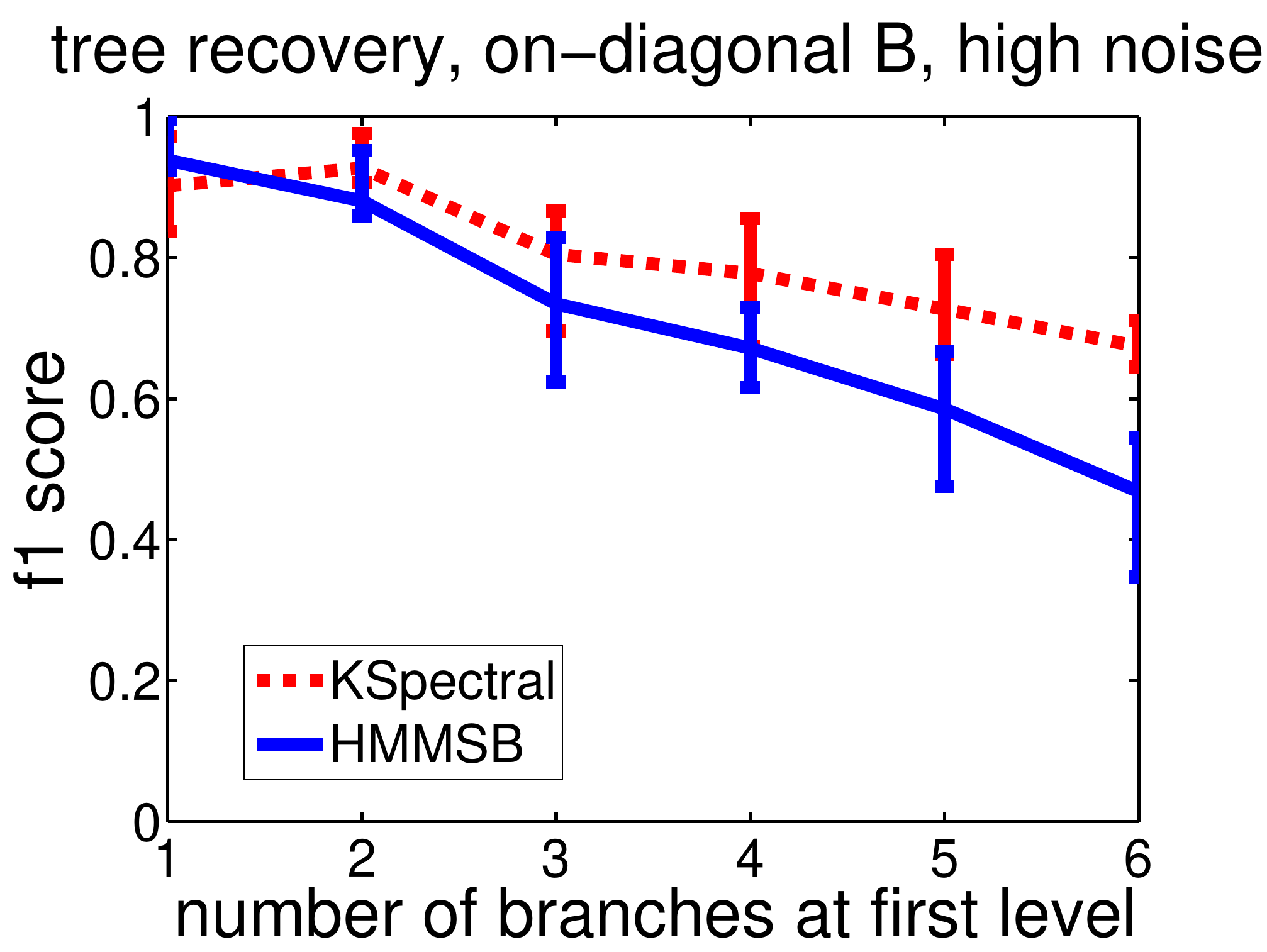}}
 \nolinebreak
 \subfigure[]{\label{fig:150offdiagonallownoise}\includegraphics[width=3.5cm]{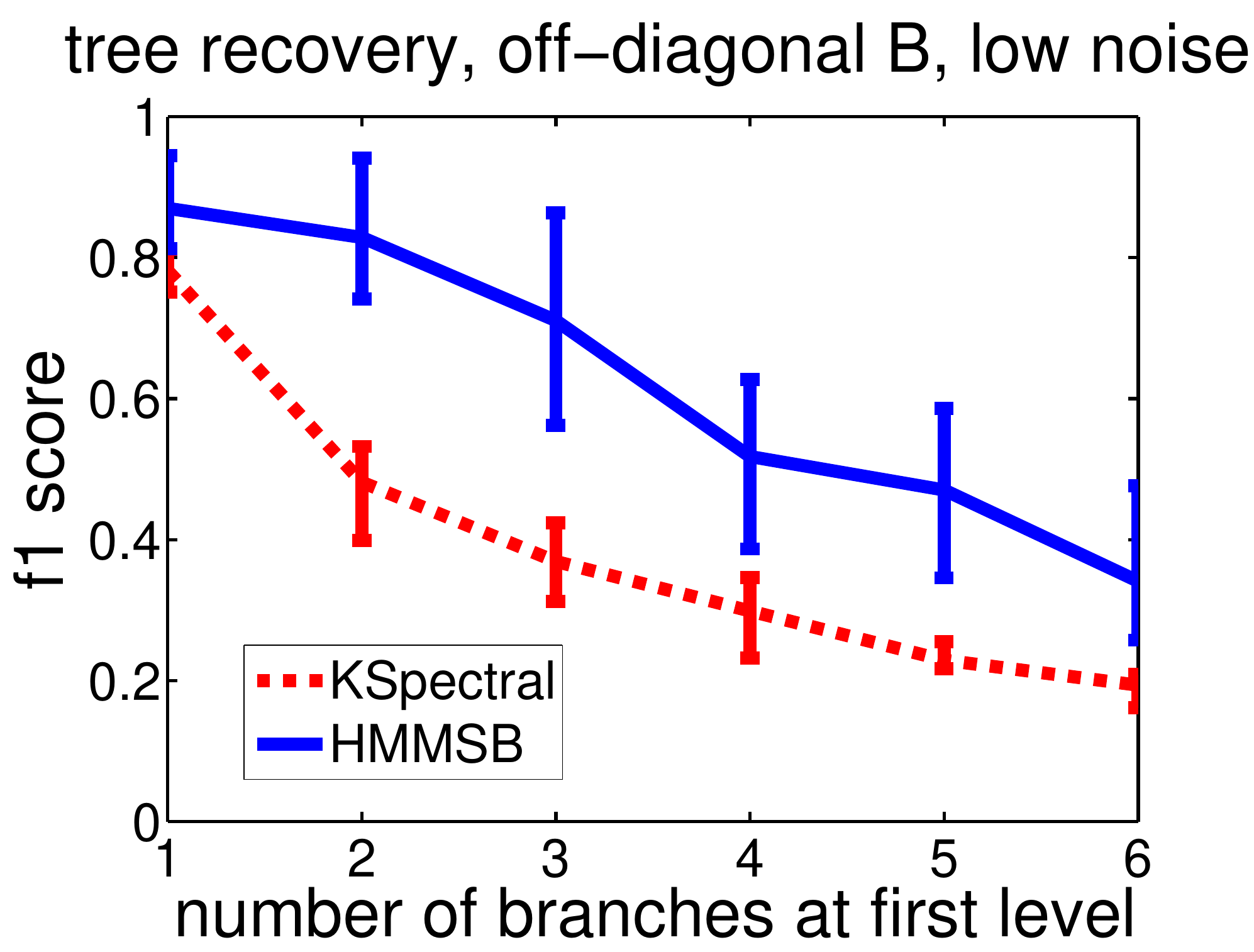}}
 \nolinebreak
 \subfigure[]{\label{fig:150offdiagonalhighnoise}\includegraphics[width=3.5cm]{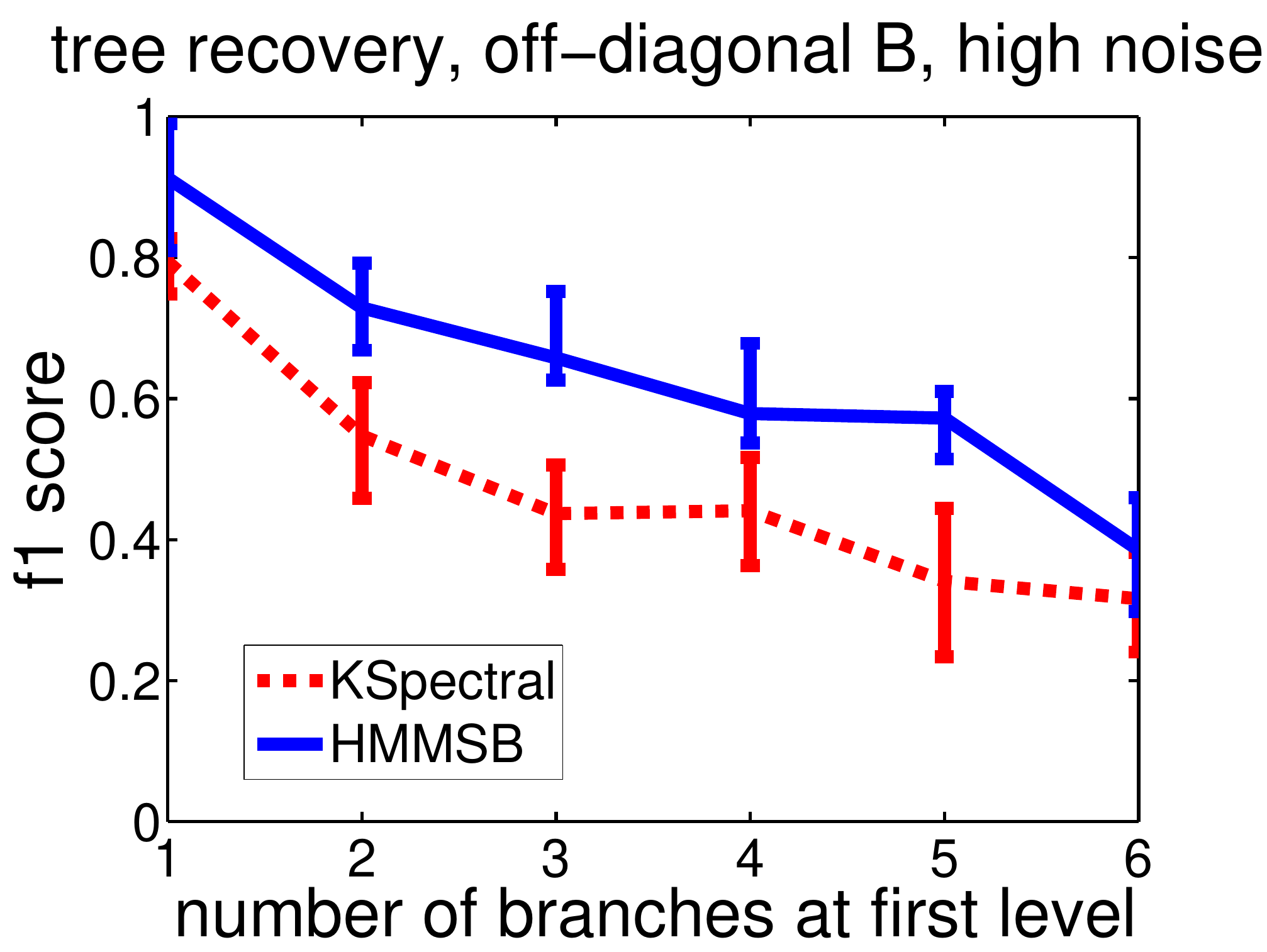}}
 \subfigure[]{\label{fig:ondiagonalgrid}\includegraphics[height=1.4cm]{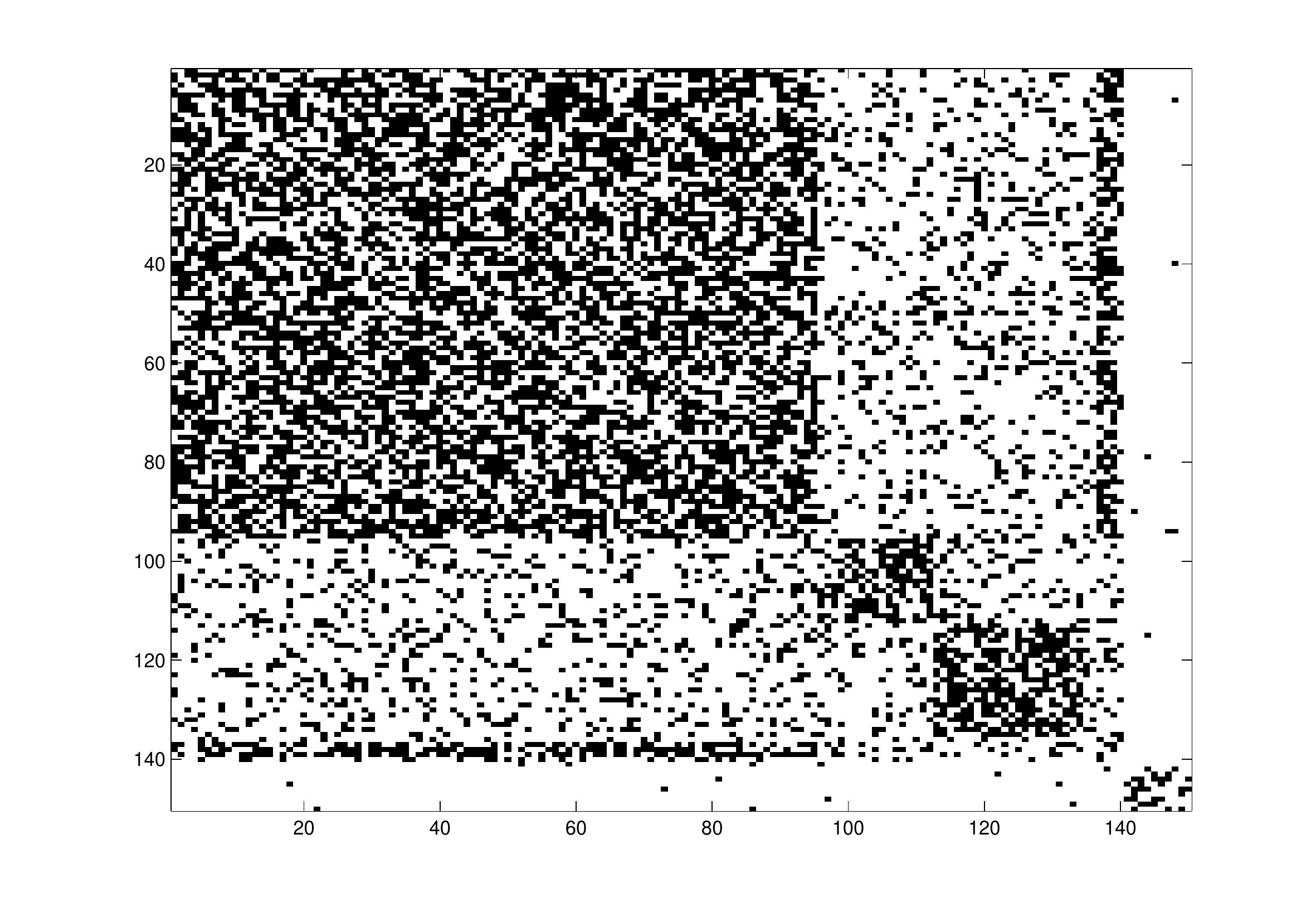}}
  \nolinebreak
 \subfigure[]{\label{fig:ondiagonaloriginal}\includegraphics[height=1.4cm]{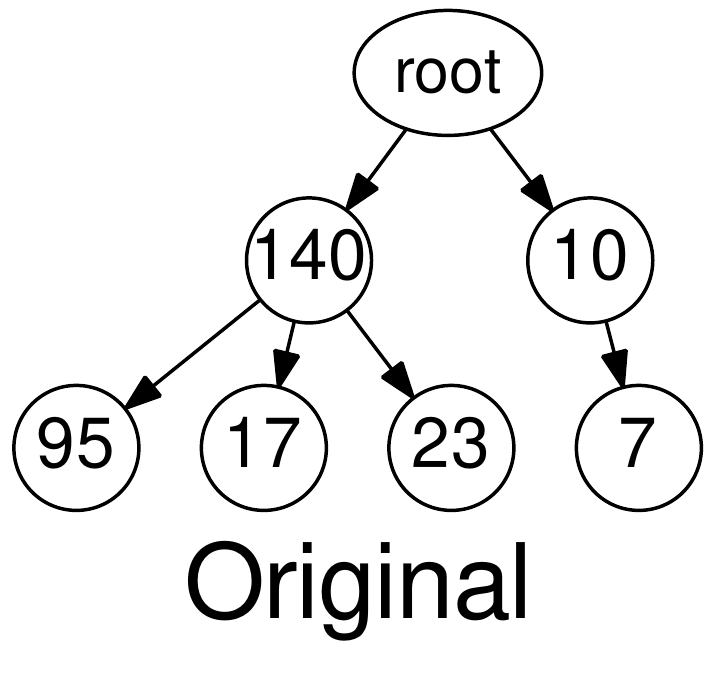}}
  \nolinebreak
 \subfigure[]{\label{fig:ondiagonalhmmsb}\includegraphics[height=1.4cm]{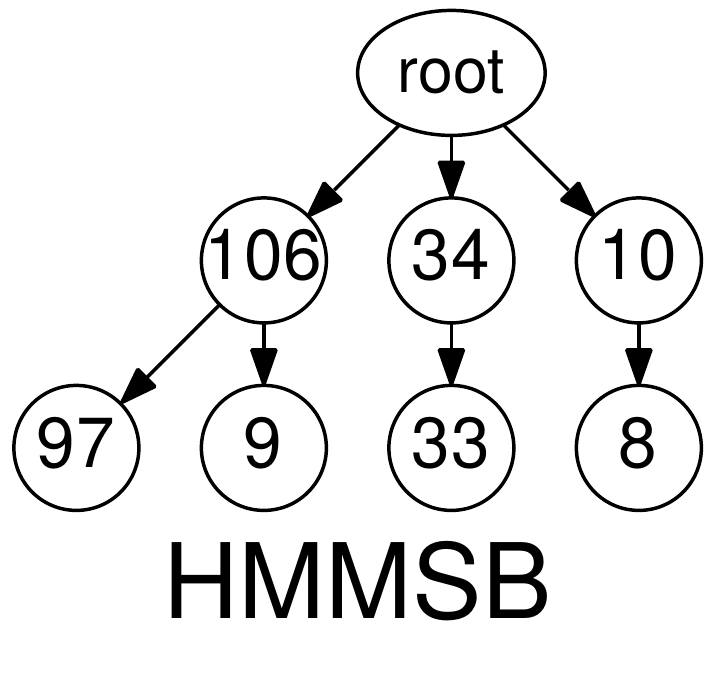}}
  \nolinebreak
 \subfigure[]{\label{fig:ondiagonalspectral}\includegraphics[height=1.4cm]{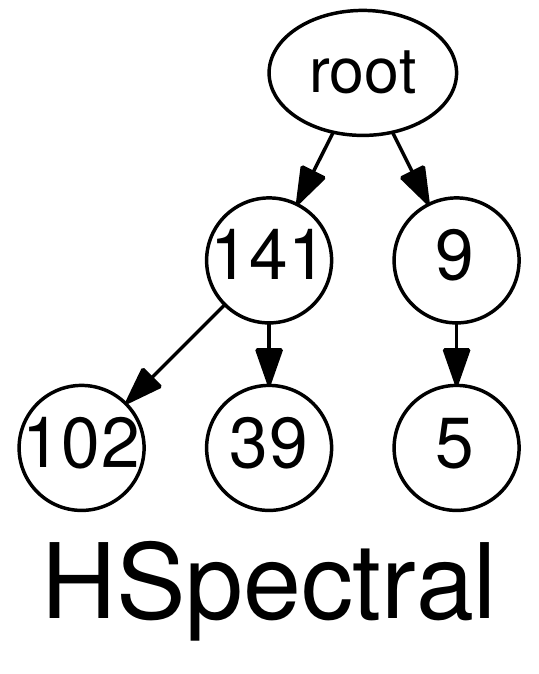}}
  \nolinebreak
 \subfigure[]{\label{fig:offdiagonalgrid}\includegraphics[height=1.4cm]{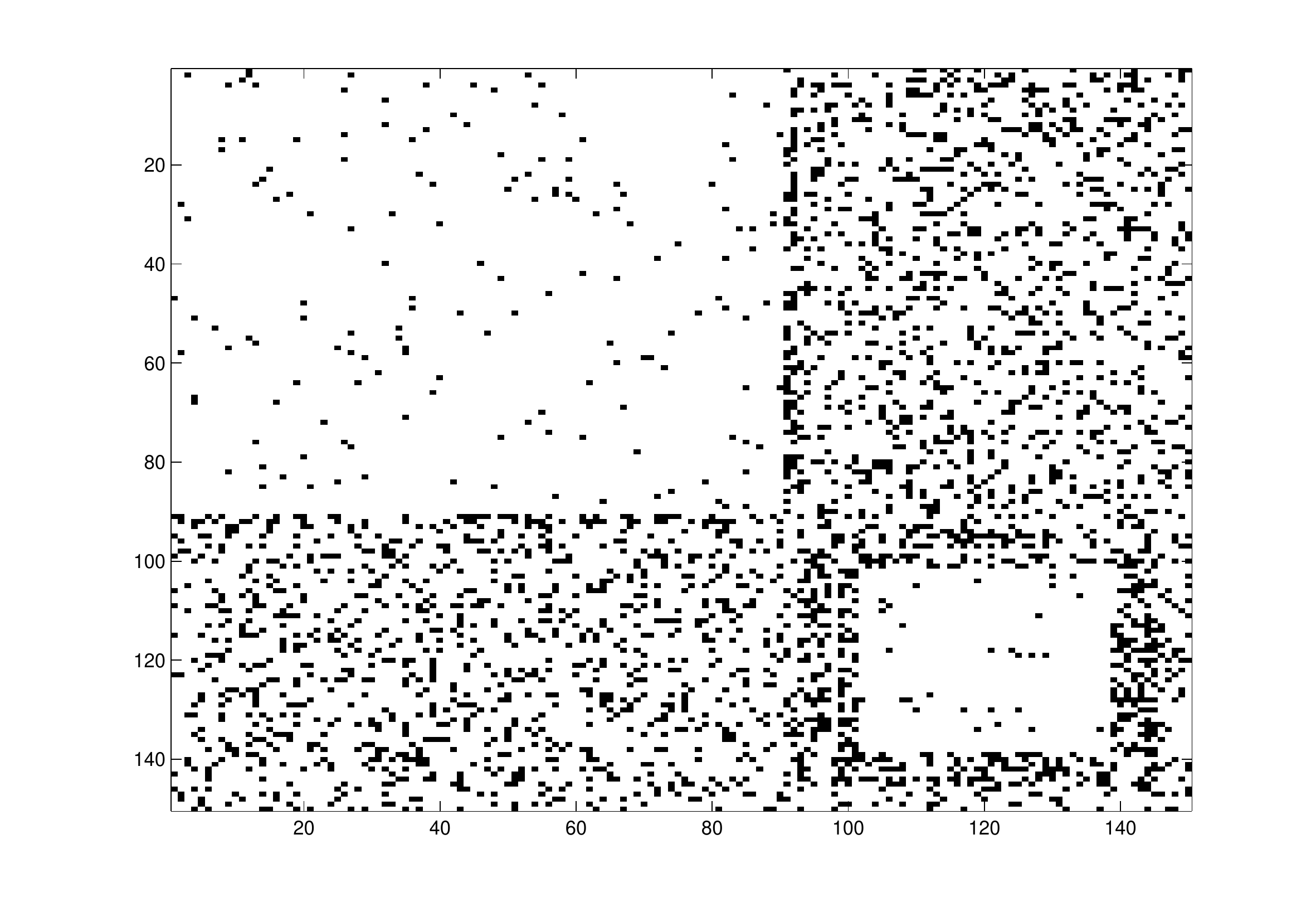}}
  \nolinebreak
 \subfigure[]{\label{fig:offdiagonaloriginal}\includegraphics[height=1.4cm]{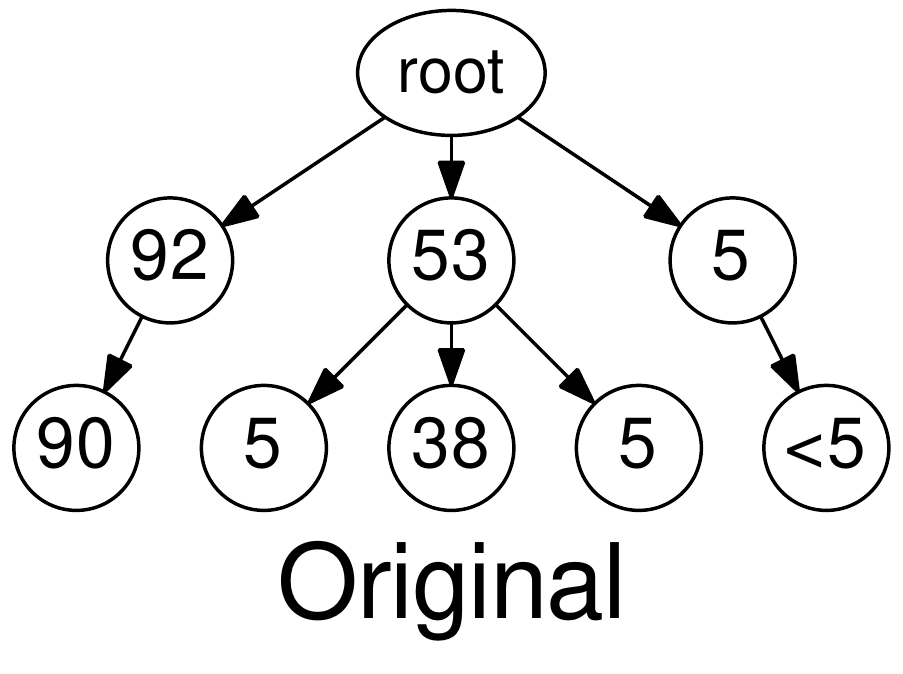}}
  \nolinebreak
 \subfigure[]{\label{fig:offdiagonalhmmsb}\includegraphics[height=1.4cm]{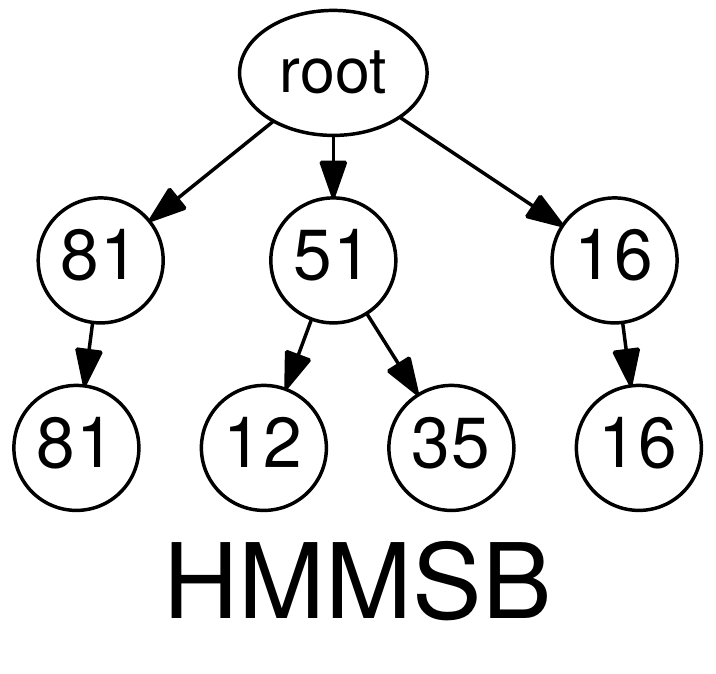}}
  \nolinebreak
 \subfigure[]{\label{fig:offdiagonalspectral}\includegraphics[height=1.4cm]{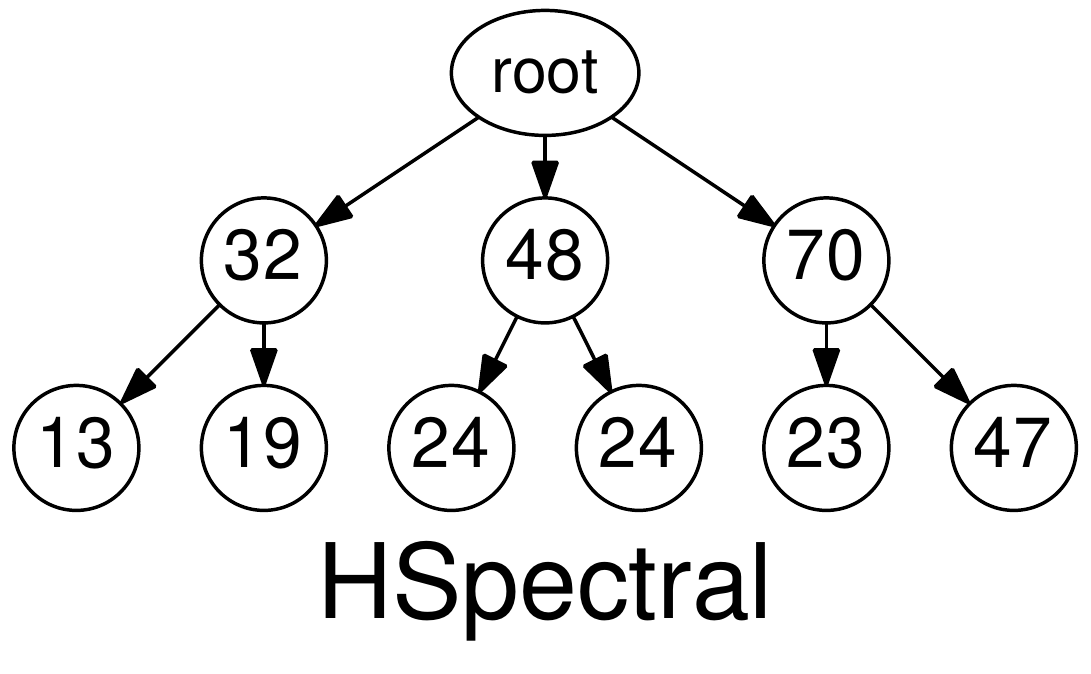}}
\end{center}

\caption{\footnotesize Simulation Results. Figures~\ref{fig:150ondiagonallownoise}, \ref{fig:150ondiagonalhighnoise}, \ref{fig:150offdiagonallownoise}, and \ref{fig:150offdiagonalhighnoise} show quantitative results. Figures~\ref{fig:ondiagonalgrid}, \ref{fig:ondiagonaloriginal}, \ref{fig:ondiagonalhmmsb} \ref{fig:ondiagonalspectral} illustrate results for one on-diagonal network, and Figures~\ref{fig:offdiagonalgrid}, \ref{fig:offdiagonaloriginal}, \ref{fig:offdiagonalhmmsb} , \ref{fig:offdiagonalspectral} illustrate results for one off-diagonal network. \ref{fig:ondiagonalgrid} and \ref{fig:offdiagonalgrid} are the original networks for these two cases (black indicates edge). The numbers inside hierarchy nodes are actor counts (nodes of size $<5$ are not shown). See text for details.
}
\label{fig:simulationresults}

\end{figure}
Figure~\ref{fig:simulationresults} illustrates the results as a function of the number of branches at the first level of the generated tree. The ``number of branches at level 1" refers to number of branches of size $\geq 5$ (since there are often branches of size 1 or 2). For fairness, the ``correct" number of level-1 branches given to spectral clustering is also the number of branches of size $\geq 5$.

As one can see, in Figure~\ref{fig:150ondiagonallownoise} and Figure~\ref{fig:150ondiagonalhighnoise}, when $\Bv$ is strongly on-diagonal, our algorithm performs well, but a little worse than HSpectral (since HSpectral is given the number of level-1 branches). A specific example is shown in Figures~\ref{fig:ondiagonalgrid}, \ref{fig:ondiagonaloriginal}, \ref{fig:ondiagonalhmmsb}, \ref{fig:ondiagonalspectral} where both models perform reasonably well. However, when $\Bv$ is strongly off-diagonal (implying that actors tend to interact with those in different communities and rather than within their own community), HSpectral performs poorly. Our method still gives good results (Figure~\ref{fig:150offdiagonallownoise}, Figure~\ref{fig:150offdiagonalhighnoise}). An example is shown in Figures~\ref{fig:offdiagonalgrid}, \ref{fig:offdiagonaloriginal}, \ref{fig:offdiagonalhmmsb}, \ref{fig:offdiagonalspectral} where our model performs accurately while HSpectral essentially divides the actors randomly and performs poorly. Thus HMMSB successfully models a variety of community interactions that traditional clustering methods cannot. It can also recover the actor-specific interaction levels for a richer network analysis.

\section{Held-out Evaluation}

\begin{wrapfigure}{r}{0.3\textwidth}

	\begin{center}
		\includegraphics[width=0.3\textwidth]{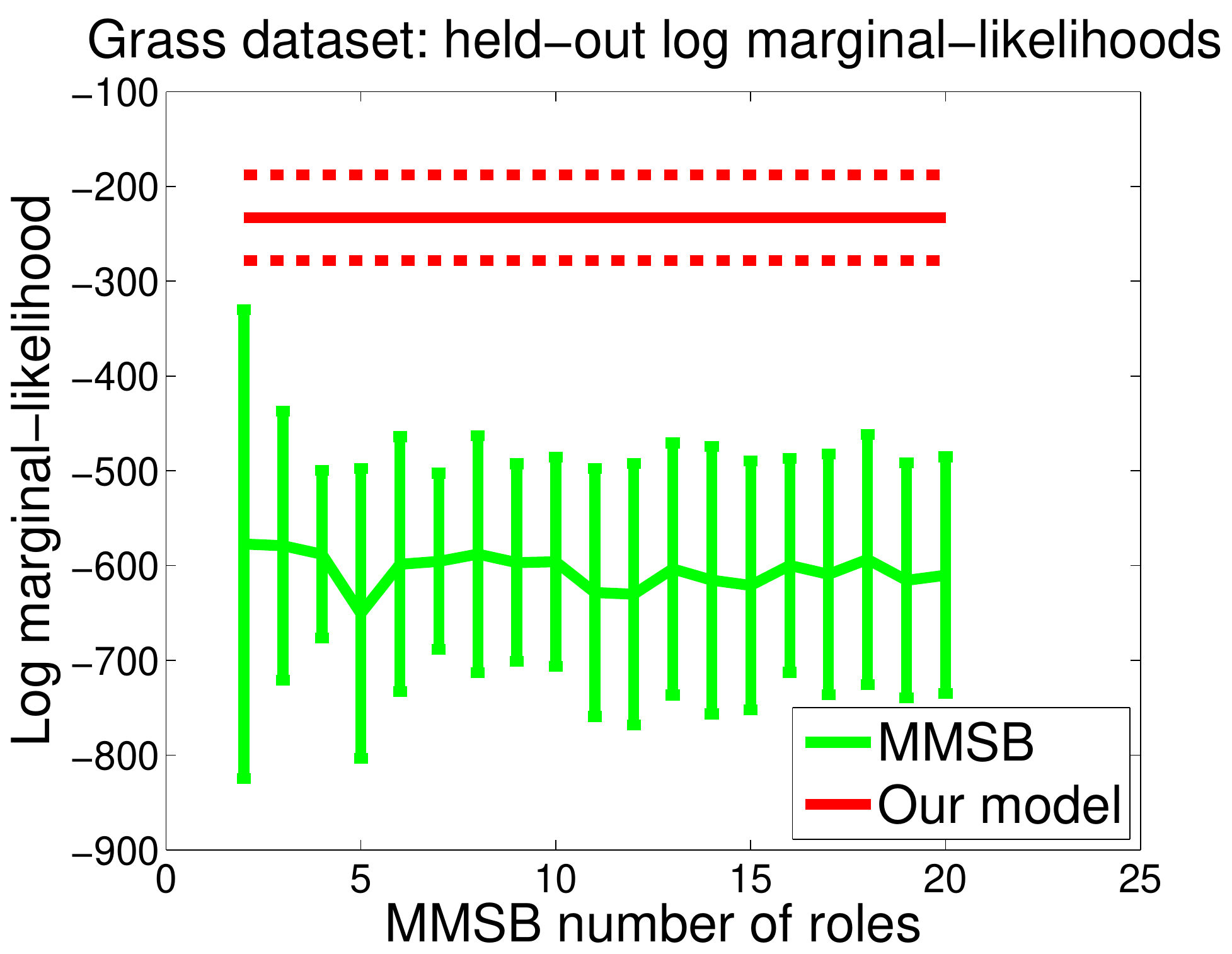} \\
		\includegraphics[width=0.31\textwidth]{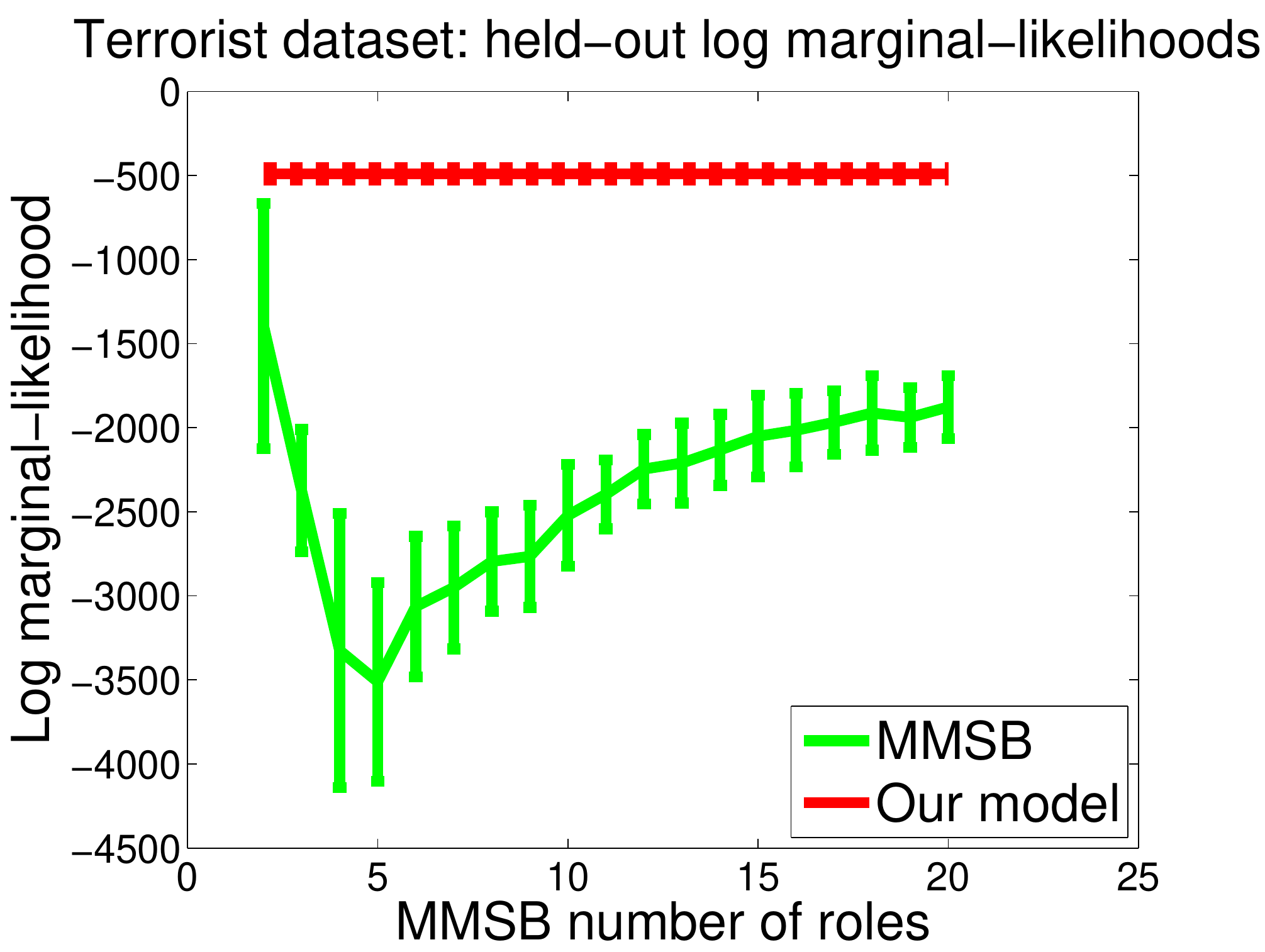}
	\end{center}

	\caption{\footnotesize
		Marginal likelihood for MMSB and hMMSB (top: grass, bottom: terrorist).
		Dotted lines show hMMSB's error bars.
		}
	\label{fig:heldout_likelihood}

\end{wrapfigure}

We compare our algorithm to MMSB \cite{airoldi2008mixed} on two real-world datasets, a 75-species food web of grass-feeding wasps \cite{dawah1995grass,clauset2008hierarchical}, and the September 11th, 2001 hijacker terrorist network \cite{krebs2002terrorist,clauset2008hierarchical}. Our choices reflect two common modes of interaction seen in real-world network data: edges in the food web denote predator-prey relationships, while edges in the terrorist network reflect social cohesion. The food web could be represented as a hierarchy where different branches reflect different trophic levels (e.g. parasite, predator or prey), while the terrorist network could be interpreted as an organization chart.

For each dataset, we generated 5 sets of training and test subgraphs, each obtained by randomly partitioning the actors into two equal sets. For each training subgraph, we performed a gridsearch over the $\Bv$ parameters $(\lambda_{1},\lambda_{2})\in\{.1,.3,.5,.7,.9\}^2$ according to the log marginal likelihood. The remaining parameters were fixed to $\gamma=1,m=\pi=0.5$ and $K=2$. We then computed the log marginal likelihood on the corresponding 5 test subgraphs, and averaged them to obtain hMMSB's average held-out likelihood. The procedure for MMSB was similar, except that we used 100 random restarts of the MMSB variational EM algorithm \cite{airoldi2008mixed} on the training subgraphs. MMSB also requires the number of latent roles $R$ as a tuning parameter, so we repeated the experiment for each $2\le R\le20$. For either algorithm, log marginal likelihoods were estimated using 10,000 iterations of importance sampling.

The results are shown in Figure \ref{fig:heldout_likelihood}.
On either dataset, our model's held-out likelihood is superior to MMSB for all $R$. Notably, MMSB's likelihood peaks on both datasets at $R=2$, but selecting so few roles will lead to an extremely coarse network analysis. In contrast, our model automatically recovers a suitable level of hierarchical complexity and enables rich interpretations of the data --- as we shall demonstrate next. We note that the ``annotated hierarchies" model \cite{roy2007learning} and infinite stochastic blockmodel \cite{kemp:pnas:2008} are good candidates for evaluation, but the authors did not make code available for computing marginal likelihoods, so we do not include those models in our evaluation.

\section{Real-data Qualitative Analysis}

\subsection{Grass-feeding Wasp Parasitoids Food Web}
We now use hMMSB to interpret real-world social networks, beginning with the grass dataset from earlier. 
We ran our Gibbs sampler on the full network to infer the community hierarchy structure and actor latent
community MMs. The model parameters were chosen on the full network via grid search over $(\lambda_{1},\lambda_{2})\in\{.1,.3,.5,.7,.9\}^2$,
according to the log marginal likelihood (estimated using 10,000 importance samples). The remaining parameters were
fixed to $\gamma=1,m=0.5,\pi=0.5$ and $K=2$. We ran our Gibbs sampler on the optimal parameters $\lambda_{1}=0.1,\lambda_{2}=0.5$
for 10,000 iterations of burn-in, and took 100 samples with a lag time of 5 iterations. Convergence was determined from a plateauing
log complete likelihood plot.

The Gibbs samples represent a posterior distribution over paths $c_i$. In order to represent the ``average" of this posterior, we generated
a consensus sample by counting the number of times each pair of actors shared the same community hierarchy position, over all samples.
Actors that shared positions in $>50\%$ of all samples were assigned to the same path in the consensus. For levels $z_{i\rightarrow j}$
and $z_{i\leftarrow j}$, we simply took the mode over all samples. In a final post-processing step to reduce visual clutter, we merged bottom-level (i.e. level-2) communities with $\le 5$ actors into one community under the same parent.

\begin{figure}[!t]
	\begin{center}
	\includegraphics[width=9cm]{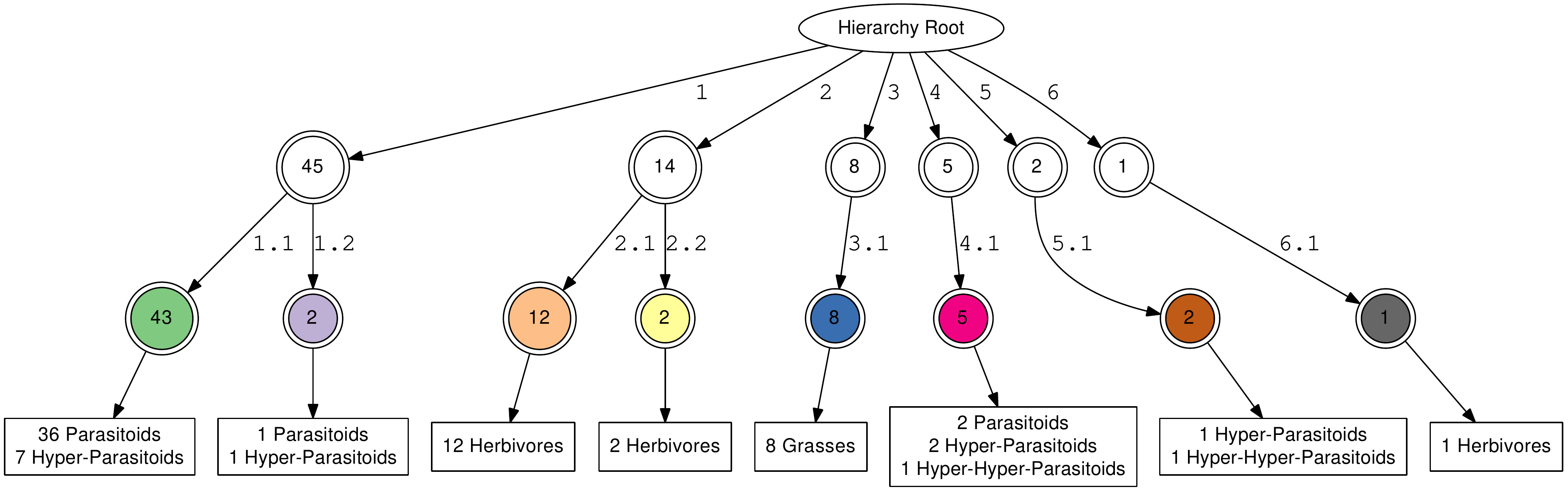} \nolinebreak
	\includegraphics[width=5cm,height=3cm]{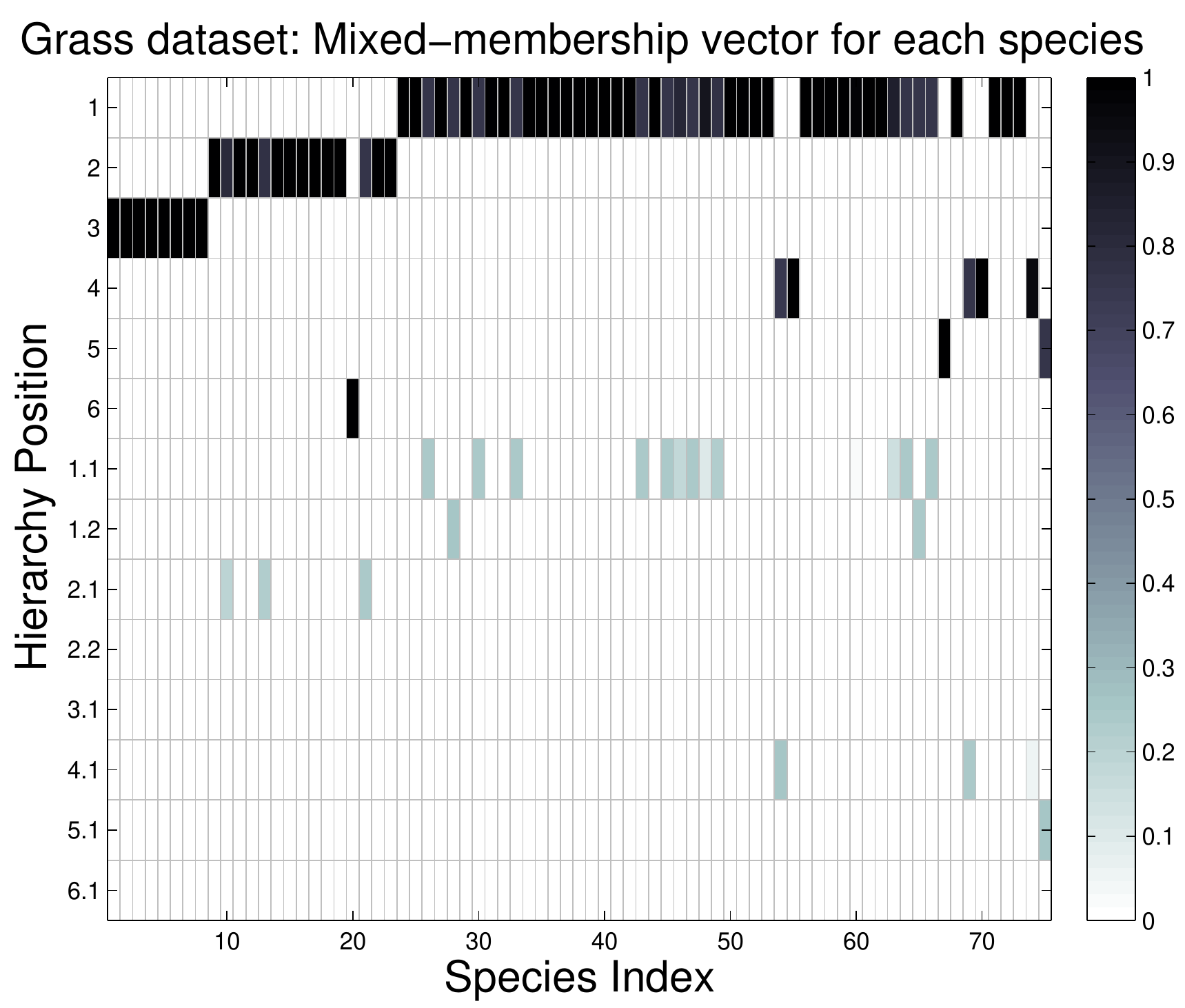} \\
	\includegraphics[width=14cm]{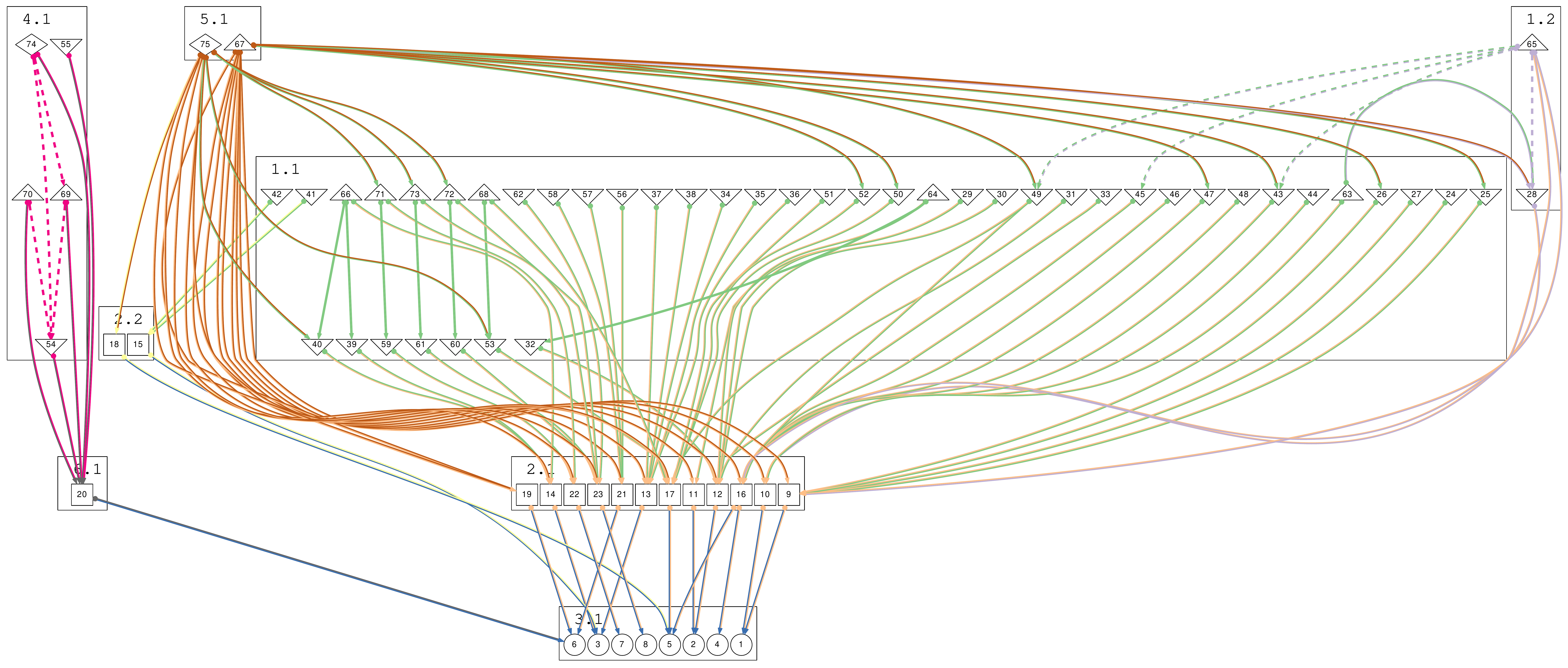}
	\begin{picture}(\textwidth,0)(0,0)
		\put(0.7\textwidth,0.02\textwidth){\includegraphics[width=0.3\textwidth]{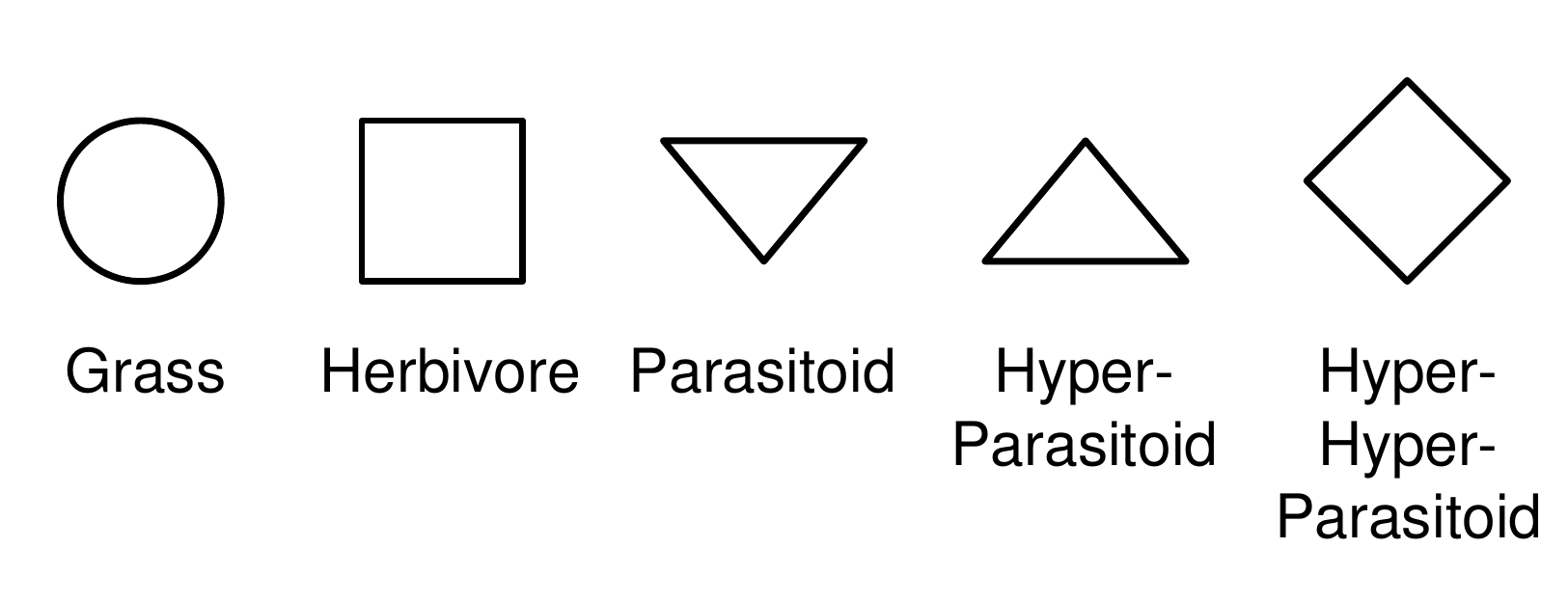}}
	\end{picture}
	\end{center}
	\caption{\footnotesize
	Grass network:
	{\bf Top Left:} Inferred hierarchy of communities, with community trophic level counts at the bottom.
	{\bf Top Right:} Community-mixed-membership vectors of each actor.
	{\bf Bottom:} Original network. Edges show interacting communities (edge head/tail
	colors match assumed hierarchy underlying the interactions) and interaction level (1 = solid, 2 = dashed) inferred by hMMSB.
	Node shapes represent annotated trophic levels (see legend in bottom right).
	}
	\label{fig:grass_qualitative}
\end{figure}

The inferred community hierarchy and MM vectors from our Gibbs sampler are reported in Figure \ref{fig:grass_qualitative}.
We also show the original network, where links $E_{ij}=1$ have been augmented with their
communities and interaction levels (missing links $E_{ij}=0$ are not shown).
The dataset contains trophic level annotations, shown in the hierarchy as counts
and in the network as node shapes.

We see that community 3 contains all grass species, 2 contains most herbivores, and 1 contains
most parasitoids; in contrast, an assortative clustering algorithm would not discover these multi-partite divisions.
The ``outlier" communities are still more interesting --- the herbivore in community 6 is the sole
prey of the parasitoids in community 4; hMMSB has separated this sub-web from the main food-web. Moreover,
the herbivores in community 2.2 are the sole prey of species 42 and 41 in community 1.1.
We also observe that community 5's two apex parasitoids have the largest and 2nd-largest range of prey species,
while community 1.2 contains the apex parasitoid with the third-largest range. As to why these communities are separated,
we note that the parasitoid in sub-community 1.2 preys on only two herbivores, compared to at least six herbivores
for either parasitoid in community 5.

The community MM vectors in Figure \ref{fig:grass_qualitative} show
the frequency at which each species identifies itself with a particular community.
Most species identify at the super-community level, though some occasionally identify at the
sub-community level. In our model, level-2 interactions occur only within super-communities, hence they
account for fine-grained, within-community interactions. For example, the within-community links in community 4,
as well as the links from species 65 in sub-community 1.2 to other members of community 1, are all level-2 interactions.
Although we have not shown interaction levels for missing links, the latter are sometimes accounted for by level-2 interactions
(e.g. in community 1).

\subsection{High-energy Physics Citation Network}

\begin{wrapfigure}{r}{0.35\textwidth}

	\begin{center}
		\includegraphics[width=0.35\textwidth]{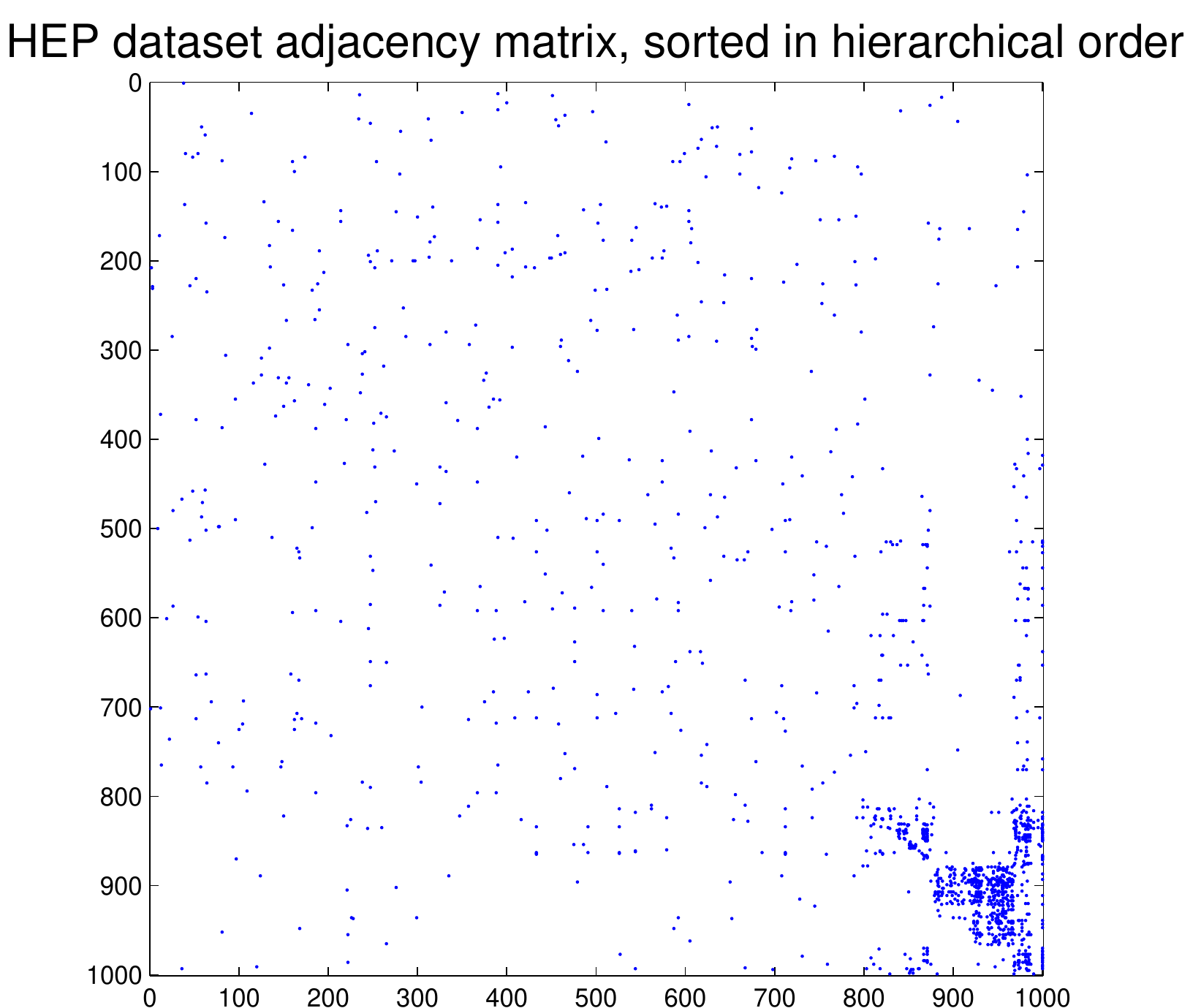}
	\end{center}

	\caption{\footnotesize HEP network: Adjacency matrix, permuted according to the
		communities in Figure \ref{fig:hep_hierarchy}.}
	\label{fig:hep_adj_matrix}

\end{wrapfigure}

Finally, we consider a 1,000-paper subgraph of the arXiv high-energy physics citation network \cite{kdd03},
which we constructed by subsampling papers involved in citations from Jan 2002 through May 2003.

We applied the same parameter selection, convergence criteria, and post-processing as the previous dataset, and the optimal parameters
were $(\lambda_{1}=0.7,\lambda_{2}=0.5)$. Each of the 25 parameter combinations required less than 6 hours to
test on a single processor core. We ran our Gibbs sampler for 10,000 iterations of burn-in, and took 10 samples with a
lag time of 50 iterations. The entire Gibbs sampling procedure completed in just under 23 hours on a single processor core.

\begin{figure}[!t]
	\begin{center}
	\includegraphics[width=15cm]{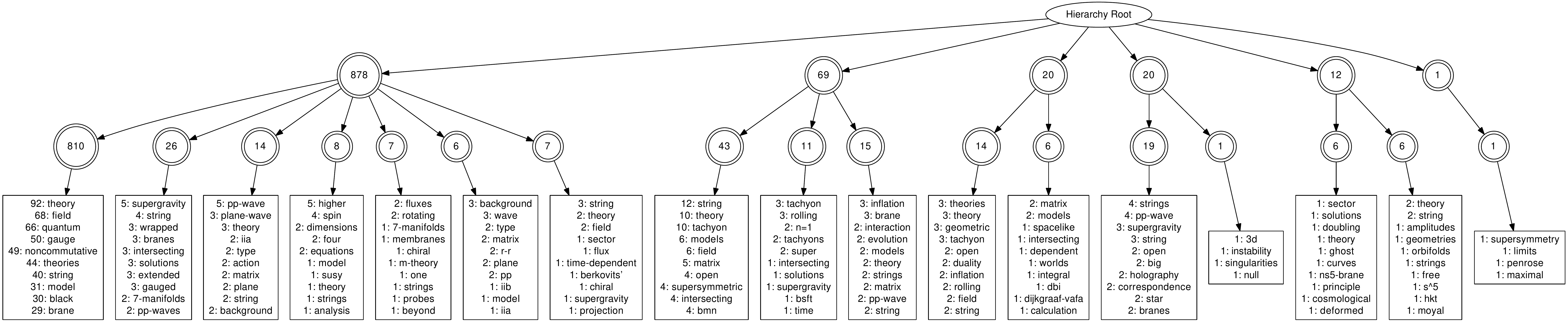}
	\end{center}
	\caption{\footnotesize
	HEP network: Inferred hierarchy of communities, with the most frequent title keywords at the bottom.
	Community positions (circles) show the number of papers.
	}
	\label{fig:hep_hierarchy}
\end{figure}

The inferred community hierarchy is shown in Figure \ref{fig:hep_hierarchy}, where 
each sub-community has been annotated with its papers' most frequent title words\footnote{While this output is reminiscent of topic models,
we stress that hMMSB is {\it not} a topic model --- the actor paths are learnt solely from the citation network, independent of the
paper contents.}. We also show the adjacency matrix in Figure \ref{fig:hep_adj_matrix}, permuted to match
the order of inferred communities.
We observe that the 810-paper sub-community has a sparse citation pattern, implying that its
papers are not specific to a particular research topic. This is confirmed by the top 3 keywords: ``theory", ``field" and ``quantum", which are
general to physics research. The other sub-communities from the same super-community are more focused, with top keywords
like ``supergravity", ``string" and ``pp-wave" being more specific physical concepts. This is also reflected in the
adjacency matrix, which is denser among these sub-communities. The remaining super-communities form a very dense sub-network that is
mostly separated from the rest of the graph, suggesting that these papers might come from a specific community of researchers,
working on a narrow set of research topics. In particular, three of the sub-communities involve the title keyword ``tachyon", which is
absent from the large super-community. To this point, we have only scratched the surface of the citation network --- a deeper analysis
would involve the abstracts/contents of the papers in the hMMSB-inferred communities.

\section{Conclusion}

We have developed a tree-structured hierarchical Mixed-Membership Stochastic Blockmodel (hMMSB) that models social networks in terms of the multiple, hierarchical community memberships that actors undertake during interactions. Our model automatically infers the number of sub-communities in each community while simultaneously recovering the community-mixed-memberships of every actor, setting it apart from hierarchy-discovering methods that are restricted to binary hierarchies and/or single-role-memberships for actors. Moreover, hMMSB is expressive enough to account for non-diagonal community-compatibility matrices, as we have demonstrated through our simulation and grass dataset experiments. On real networks, we show that hMMSB recovers intuitive, mixed-membership community organization. Finally, our collapsed Gibbs sampler efficiently scales to medium-sized datasets of around 1,000 actors, completing inference in a single day on a single processor core.

\section*{Acknowledgements}
This paper is based on work supported by NSF IIS-0713379, AFOSR FA9550010247, ONR N000140910758, DARPA NBCH1080007, NIH 1R01GM093156, and an 
Alfred P.  Sloan Research Fellowship to Eric P. Xing.
Qirong Ho is supported by a graduate fellowship from the Agency for Science, Technology And Research, Singapore.

\bibliography{hmmsbbib} \bibliographystyle{plain}

\end{document}